\def\arxivversion{1}

\ifdefined\standalonesupplement
\documentclass[10pt,twocolumn,letterpaper]{article}
\else
\documentclass[10pt,letterpaper,twocolumn]{article}
\fi
\newif\ifsupplement
\ifdefined\standalonesupplement
\supplementtrue
\else
\supplementfalse
\fi
\usepackage[T1]{fontenc}
\ifdefined\reviewversion
\usepackage[review]{cvpr}
\else
\usepackage[pagenumbers]{cvpr}
\fi
\usepackage{times}
\def\confName{CVPR}
\def\confYear{2027}

\usepackage{amsmath,amsfonts,bm}

\def\eqref#1{equation~\ref{#1}}

\def\1{\bm{1}}

\DeclareMathAlphabet{\mathsfit}{\encodingdefault}{\sfdefault}{m}{sl}
\SetMathAlphabet{\mathsfit}{bold}{\encodingdefault}{\sfdefault}{bx}{n}

\definecolor{cvprblue}{rgb}{0.21,0.49,0.74}
\ifdefined\arxivversion
\usepackage[breaklinks,colorlinks,allcolors=cvprblue]{hyperref}
\else
\usepackage[pagebackref,breaklinks,colorlinks,allcolors=cvprblue]{hyperref}
\fi
\usepackage{url}
\usepackage{graphicx}
\usepackage{booktabs}
\usepackage{array}
\usepackage{microtype}
\usepackage{xcolor}
\usepackage{colortbl}
\usepackage{amsmath}
\usepackage{amssymb}
\DeclareMathSizes{7.5}{7}{5}{5}
\usepackage{multirow}
\usepackage{placeins}
\usepackage{float}
\usepackage{needspace}
\usepackage{xspace}
\usepackage{wrapfig}
\usepackage{tabularx}
\usepackage{subcaption}
\usepackage{multicol}

\usepackage{tikz}
\usetikzlibrary{arrows.meta}

\title{SegDis: Foreground Segmentation Followed by Type Discrimination for Salient and Camouflaged Object Detection}

\author{\small Dingning Luo$^{1\ast}$ \quad Shuo Dou$^{1\ast}$ \quad Tong Jiang$^{1\ast}$ \quad
Zhenhua Duan$^{1}$ \quad Jiachen Zhu$^{1}$ \quad Xianfeng Li$^{1\dagger}$\\[2pt]
$^{1}$Macau University of Science and Technology}

\newcommand{\method}{\textsc{SegDis}\xspace}
\newcommand{\flr}{\textsc{FLR}\xspace}

\newcommand{\modelcite}[2]{\hyperlink{cite.#2}{#1}~\citep{#2}}

\definecolor{tblheader}{HTML}{F2F2F2}
\definecolor{tblgroup}{HTML}{FFFFFF}
\definecolor{tblours}{HTML}{E6EFF6}
\definecolor{tbljoint}{HTML}{F7FAFC}
\definecolor{tblpanel}{HTML}{FFFFFF}
\definecolor{tblaccent}{HTML}{205B7E}
\newcommand{\tblhead}[1]{\textbf{\boldmath #1}}
\newcolumntype{L}[1]{>{\raggedright\arraybackslash}p{#1}}
\newcolumntype{N}[1]{>{\raggedleft\arraybackslash}p{#1}}

\newsavebox{\segdistablebox}
\newcommand{\fitsegdistable}{%
  \ifdim\wd\segdistablebox>\columnwidth
    \resizebox{\columnwidth}{!}{\usebox{\segdistablebox}}%
  \else
    \usebox{\segdistablebox}%
  \fi}
\newsavebox{\segdisimagebox}
\newcommand{\segdismontage}[4]{%
  \sbox{\segdisimagebox}{\includegraphics[width=.88\linewidth,
    trim=#2bp 0bp 0bp #3bp,clip]{#1}}%
  \hbox to\linewidth{%
    \hbox to.12\linewidth{}%
    \hbox to.1466\linewidth{\hss\fontsize{6.5}{7.4}\selectfont Input\hss}%
    \hbox to.1466\linewidth{\hss\fontsize{6.5}{7.4}\selectfont\shortstack{Ground\\truth}\hss}%
    \hbox to.1466\linewidth{\hss\fontsize{6.5}{7.4}\selectfont\shortstack{Unified\\FG}\hss}%
    \hbox to.1466\linewidth{\hss\fontsize{6.5}{7.4}\selectfont SOD\hss}%
    \hbox to.1466\linewidth{\hss\fontsize{6.5}{7.4}\selectfont COD\hss}%
    \hbox to.1466\linewidth{\hss\fontsize{6.5}{7.4}\selectfont\shortstack{Final\\output}\hss}}%
  \vspace{2pt}\par\nointerlineskip
  \hbox to\linewidth{%
    \vbox to\ht\segdisimagebox{\vfil
      \foreach \caseid [count=\caseindex] in {#4}{%
        \ifnum\caseindex>1\vfil\fi
        \hbox to.12\linewidth{\hss\fontsize{6.3}{7.3}\selectfont\caseid\hss}\vfil}}%
    \usebox{\segdisimagebox}}%
}

\renewcommand{\arraystretch}{1.10}

\begin{document}
\ifsupplement
\setcounter{dbltopnumber}{4}
\maketitlesupplementary
\appendix
\renewcommand{\thetable}{S\arabic{table}}
\renewcommand{\thefigure}{S\arabic{figure}}
\renewcommand{\theequation}{S\arabic{equation}}
\renewcommand{\thepage}{S\arabic{page}}
\setlength{\parskip}{0pt}
\setlength{\abovedisplayskip}{4pt}
\setlength{\belowdisplayskip}{4pt}

This supplement provides the complete foreground benchmarks, the full soft
metric tables for the three training modes, additional type-allocation cases,
and implementation details that are omitted from the main paper for space.
All experiments use the same model definitions and evaluation protocols as the
main paper. External baseline entries are transcribed from the cited papers;
they are included for context and are not reimplemented here.

\section{Additional foreground benchmark results}
\label{sec:supp-foreground}

The main paper reports compact SOD, COD, and joint-training comparisons in
Tables~3--5. Tables~\ref{tab:exp-sod-comparison} and
\ref{tab:exp-cod-comparison} give the corresponding complete benchmark
tables. ``Spec.'' denotes task-specific training, ``Joint'' denotes one model
trained on paired SOD/COD data, and the parameter column reports total model
parameters in millions. The SegDis rows use the PVT-v2-B2 backbone and the
same binary foreground probability used for the main-paper comparisons.

The SOD-trained model is strongest on the SOD benchmarks, whereas the
COD-trained model is strongest on the COD benchmarks. Joint training preserves
strong performance in both regimes: it reaches $S_\alpha=.9235$ on DUTS-TE
and $S_\alpha=.8693$ on COD10K. This separation clarifies the role of the
foreground pathway: the same FLR design can be evaluated under specialized
or paired supervision, while the typed USC12K experiment adds a separate
RTA decision for SOD/COD allocation.

\begin{table*}[t]\noindent\begin{minipage}{\textwidth}
\captionof{table}{Conventional SOD comparison. Spec./Joint denote task-specific/joint SOD/COD training. Params: total millions; external rows are transcribed from the cited primary papers; best values in each family are bolded.}
\label{tab:exp-sod-comparison}
\centering\footnotesize
\setlength{\tabcolsep}{2.0pt}\renewcommand{\arraystretch}{1.06}
\begin{tabular}{@{}L{.06\textwidth}L{.21\textwidth}L{.135\textwidth}N{.065\textwidth}*{10}{N{.0415\textwidth}}@{}}
\toprule

& & & & \multicolumn{2}{c}{\textbf{DUTS-TE}} & \multicolumn{2}{c}{\textbf{ECSSD}} & \multicolumn{2}{c}{\textbf{HKU-IS}} & \multicolumn{2}{c}{\textbf{DUT-OMRON}} & \multicolumn{2}{c}{\textbf{PASCAL-S}} \\
\cmidrule(lr){5-6}\cmidrule(lr){7-8}\cmidrule(lr){9-10}\cmidrule(lr){11-12}\cmidrule(l){13-14}

\tblhead{Family} & \tblhead{Method} & \tblhead{Backbone} & \tblhead{Params (M)} & \tblhead{$S_\alpha$} & \tblhead{MAE} & \tblhead{$S_\alpha$} & \tblhead{MAE} & \tblhead{$S_\alpha$} & \tblhead{MAE} & \tblhead{$S_\alpha$} & \tblhead{MAE} & \tblhead{$S_\alpha$} & \tblhead{MAE} \\
\midrule
Spec. & \modelcite{F3Net}{wei2020f3net} & ResNet-50 & 25.54 & .8880 & .0350 & .9240 & .0330 & .9170 & .0280 & .8380 & .0530 & .8610 & .0610 \\
Spec. & \modelcite{VST}{liu2021vst} & T2T-ViT-t-14 & 44.48 & .8960 & .0370 & .9320 & .0330 & .9280 & .0290 & .8500 & .0580 & .8720 & .0610 \\
Spec. & \modelcite{ICON-S}{zhuge2022icon} & Swin-B & 92.15 & .9170 & .0250 & .9410 & .0230 & .9350 & .0220 & .8690 & .0430 & .8850 & .0480 \\
\midrule
Joint & \modelcite{SENet}{hao2025senet} & Asym. ViT & 110.20 & .9260 & .0220 & .9490 & .0200 & .9410 & .0190 & .8760 & .0400 & .8900 & .0460 \\
Joint & \modelcite{JoNet}{hao2025distribution} & SENet ViT & 110.30 & .9310 & .0200 & .9480 & .0200 & .9380 & .0190 & .8720 & .0400 & .8870 & .0460 \\
Joint & \modelcite{UJSC}{li2021ujsc} & ResNet-50 & 121.63 & .8990 & .0320 & .9330 & .0300 & .9310 & .0260 & .8500 & .0510 & .8640 & .0620 \\
\midrule
Spec. & \textbf{SegDis (SOD)} & \textbf{PVT-v2-B2} & \textbf{28.80} & \textbf{.9215} & \textbf{.0232} & .9395 & .0240 & .9342 & \textbf{.0207} & \textbf{.8694} & \textbf{.0425} & .8836 & .0494\\
\bottomrule
\end{tabular}
\end{minipage}\end{table*}
\begin{table*}[t]\noindent\begin{minipage}{\textwidth}
\captionof{table}{Conventional COD comparison. Training families and parameter notation follow Table~\ref{tab:exp-sod-comparison}.}
\label{tab:exp-cod-comparison}
\centering\footnotesize
\setlength{\tabcolsep}{2.5pt}\renewcommand{\arraystretch}{1.06}
\begin{tabular}{@{}L{.06\textwidth}L{.21\textwidth}L{.145\textwidth}N{.065\textwidth}*{6}{N{.071\textwidth}}@{}}
\toprule

& & & & \multicolumn{2}{c}{\textbf{CAMO}} & \multicolumn{2}{c}{\textbf{COD10K}} & \multicolumn{2}{c}{\textbf{NC4K}} \\
\cmidrule(lr){5-6}\cmidrule(lr){7-8}\cmidrule(l){9-10}

\tblhead{Family} & \tblhead{Method} & \tblhead{Backbone} & \tblhead{Params (M)} & \tblhead{$S_\alpha$} & \tblhead{MAE} & \tblhead{$S_\alpha$} & \tblhead{MAE} & \tblhead{$S_\alpha$} & \tblhead{MAE} \\
\midrule
Spec. & \modelcite{ZoomNet}{pang2022zoom} & ResNet-50 & 32.38 & .8200 & .0660 & .8380 & .0290 & .8530 & .0430 \\
Spec. & \modelcite{CamoFormer-P}{yin2024camoformer} & PVT-v2-B4 & 71.00 & .8720 & .0460 & .8690 & .0230 & .8920 & .0300 \\
\midrule
Joint & \modelcite{SENet}{hao2025senet} & Asym. ViT & 110.20 & .8880 & .0390 & .8650 & .0240 & .8890 & .0320 \\
Joint & \modelcite{JoNet}{hao2025distribution} & SENet ViT & 110.30 & .8890 & .0380 & .8710 & .0230 & .8950 & .0310 \\
Joint & \modelcite{UJSC}{li2021ujsc} & ResNet-50 & 121.63 & .8030 & .0760 & .8170 & .0350 & .8420 & .0470 \\
\midrule
Spec. & \textbf{SegDis (COD)} & \textbf{PVT-v2-B2} & \textbf{28.80} & .8664 & .0493 & .8673 & .0230 & .8891 & .0319\\
\bottomrule
\end{tabular}
\end{minipage}\end{table*}

\section{Additional metrics of the FLR ablation}
\par\smallskip\noindent\begin{minipage}{\columnwidth}
\centering\footnotesize\rmfamily
\captionof{table}{Metrics of the FLR ablation (rows F0--F2 of Table~7) that are
not reported in the main paper (\%). E: PVT-v2-B2 encoder, with the complete RTA
held fixed. The three rows correspond one-to-one to Table~7.}
\label{tab:supp-ablation-full}
\setlength{\tabcolsep}{3pt}\renewcommand{\arraystretch}{1.14}
\begin{tabular*}{\linewidth}{@{\extracolsep{\fill}}l*{5}{c}@{}}
\toprule
\tblhead{ID} & \tblhead{IoU-BG$\uparrow$} & \tblhead{mIoU$\uparrow$} & \tblhead{mAcc$\uparrow$} & \tblhead{Pix.\ acc$\uparrow$} & \tblhead{CSCS$\downarrow$}\\
\midrule
F0 & 96.043 & 73.762 & 83.357 & 95.805 & 6.113\\
F1 & 96.972 & 77.470 & 85.239 & 96.674 & 6.174\\
F2 & \textbf{97.086} & \textbf{78.351} & \textbf{85.945} & \textbf{96.812} & \textbf{5.752}\\
\bottomrule
\end{tabular*}
\end{minipage}\par\smallskip

\section{Complete metrics for the three training modes}
\label{sec:supp-soft-metrics}

Tables~\ref{tab:supp-full-sod}--\ref{tab:supp-full-joint} report all nine
soft metrics for the SOD-trained, COD-trained, and jointly trained SegDis on
the same nine test sets. The scores are computed from the raw unified
foreground probability $P_u$ without per-image min--max normalization.
$S_\alpha$, weighted $F_\beta$, adaptive/mean/maximum $F$ and adaptive/mean/
maximum $E$ are higher-is-better; MAE is lower-is-better.

The three tables expose the expected transfer pattern. SOD-only training
retains high scores on DUTS-TE and ECSSD but degrades on camouflage-heavy
sets such as CAMO and COD10K. COD-only training shows the complementary
pattern, with $S_\alpha=.9133$ on CHAMELEON and $S_\alpha=.8673$ on COD10K,
but substantially lower scores on several SOD sets. Paired joint training
recovers a more balanced operating point, including
$S_\alpha=.9235/.8693$ on DUTS-TE/COD10K. These results motivate reporting
both specialized and joint settings rather than treating one training mode as
representative of all foreground regimes.

\begin{table*}[t]\noindent\begin{minipage}{\textwidth}
\centering\footnotesize
\captionof{table}{Full metrics of the SOD-trained SegDis. Trained on SOD benchmarks only; all nine soft metrics on the nine test sets, $P_u$ / raw, no per-image min--max normalisation.}
\label{tab:supp-full-sod}
\setlength{\tabcolsep}{4pt}\renewcommand{\arraystretch}{1.05}
\begin{tabular*}{\textwidth}{@{\extracolsep{\fill}}l*{9}{c}@{}}
\toprule
\tblhead{Test set} & \tblhead{$S_\alpha\uparrow$} & \tblhead{$F_\beta^w\uparrow$} & \tblhead{MAE$\downarrow$} & \tblhead{adp$F\uparrow$} & \tblhead{mean$F\uparrow$} & \tblhead{max$F\uparrow$} & \tblhead{adp$E\uparrow$} & \tblhead{mean$E\uparrow$} & \tblhead{max$E\uparrow$}\\
\midrule
DUTS-TE & .9215 & .8967 & .0232 & .8620 & .8987 & .9198 & .9319 & .9519 & .9598\\
ECSSD & .9395 & .9381 & .0240 & .8963 & .9400 & .9534 & .9314 & .9605 & .9678\\
HKU-IS & .9342 & .9297 & .0207 & .9247 & .9303 & .9480 & .9651 & .9648 & .9718\\
DUT-OMRON & .8694 & .8084 & .0425 & .8038 & .8159 & .8336 & .8974 & .9001 & .9088\\
PASCAL-S & .8836 & .8572 & .0494 & .8087 & .8663 & .8851 & .8837 & .9217 & .9287\\
CAMO & .7397 & .6360 & .0963 & .6912 & .6813 & .6925 & .8117 & .7809 & .8123\\
CHAMELEON & .7332 & .5989 & .0720 & .6697 & .6473 & .6635 & .8150 & .7535 & .8169\\
COD10K & .7769 & .6476 & .0459 & .6867 & .6878 & .6966 & .8564 & .8278 & .8482\\
NC4K & .8253 & .7523 & .0529 & .7867 & .7859 & .7949 & .8799 & .8633 & .8803\\
\bottomrule
\end{tabular*}
\end{minipage}\end{table*}
\begin{table*}[t]\noindent\begin{minipage}{\textwidth}
\centering\footnotesize
\captionof{table}{Full metrics of the COD-trained SegDis. Trained on COD benchmarks only; same protocol and metrics as Table~\ref{tab:supp-full-sod}.}
\label{tab:supp-full-cod}
\setlength{\tabcolsep}{4pt}\renewcommand{\arraystretch}{1.05}
\begin{tabular*}{\textwidth}{@{\extracolsep{\fill}}l*{9}{c}@{}}
\toprule
\tblhead{Test set} & \tblhead{$S_\alpha\uparrow$} & \tblhead{$F_\beta^w\uparrow$} & \tblhead{MAE$\downarrow$} & \tblhead{adp$F\uparrow$} & \tblhead{mean$F\uparrow$} & \tblhead{max$F\uparrow$} & \tblhead{adp$E\uparrow$} & \tblhead{mean$E\uparrow$} & \tblhead{max$E\uparrow$}\\
\midrule
DUTS-TE & .8161 & .7479 & .0605 & .7916 & .7815 & .7967 & .8687 & .8433 & .8807\\
ECSSD & .7495 & .6700 & .1077 & .7408 & .7118 & .7438 & .7687 & .7452 & .7980\\
HKU-IS & .7665 & .6887 & .0834 & .7706 & .7346 & .7616 & .8089 & .7740 & .8285\\
DUT-OMRON & .6350 & .4466 & .1091 & .5475 & .4967 & .5507 & .7055 & .6381 & .7223\\
PASCAL-S & .6968 & .5931 & .1325 & .6695 & .6439 & .6970 & .7296 & .7043 & .7763\\
CAMO & .8664 & .8178 & .0493 & .8342 & .8420 & .8579 & .9176 & .9162 & .9271\\
CHAMELEON & .9133 & .8566 & .0230 & .8481 & .8670 & .8901 & .9428 & .9389 & .9519\\
COD10K & .8673 & .7800 & .0230 & .7789 & .8021 & .8261 & .9253 & .9268 & .9391\\
NC4K & .8891 & .8407 & .0319 & .8437 & .8598 & .8766 & .9306 & .9316 & .9418\\
\bottomrule
\end{tabular*}
\end{minipage}\end{table*}
\begin{table*}[t]\noindent\begin{minipage}{\textwidth}
\centering\footnotesize
\captionof{table}{Full metrics of the jointly trained SegDis. Trained on paired SOD/COD data; same protocol and metrics as Table~\ref{tab:supp-full-sod}.}
\label{tab:supp-full-joint}
\setlength{\tabcolsep}{4pt}\renewcommand{\arraystretch}{1.05}
\begin{tabular*}{\textwidth}{@{\extracolsep{\fill}}l*{9}{c}@{}}
\toprule
\tblhead{Test set} & \tblhead{$S_\alpha\uparrow$} & \tblhead{$F_\beta^w\uparrow$} & \tblhead{MAE$\downarrow$} & \tblhead{adp$F\uparrow$} & \tblhead{mean$F\uparrow$} & \tblhead{max$F\uparrow$} & \tblhead{adp$E\uparrow$} & \tblhead{mean$E\uparrow$} & \tblhead{max$E\uparrow$}\\
\midrule
DUTS-TE & .9235 & .8963 & .0230 & .8591 & .8984 & .9215 & .9320 & .9529 & .9618\\
ECSSD & .9190 & .9094 & .0349 & .8917 & .9172 & .9270 & .9205 & .9362 & .9486\\
HKU-IS & .9149 & .9006 & .0282 & .9091 & .9089 & .9220 & .9499 & .9438 & .9537\\
DUT-OMRON & .8645 & .7953 & .0440 & .7920 & .8045 & .8244 & .8927 & .8952 & .9046\\
PASCAL-S & .8693 & .8375 & .0548 & .8100 & .8523 & .8621 & .8793 & .9049 & .9156\\
CAMO & .8651 & .8169 & .0515 & .8256 & .8407 & .8575 & .9116 & .9133 & .9241\\
CHAMELEON & .9117 & .8580 & .0227 & .8497 & .8663 & .8885 & .9367 & .9416 & .9558\\
COD10K & .8693 & .7855 & .0232 & .7833 & .8073 & .8321 & .9251 & .9279 & .9397\\
NC4K & .8904 & .8424 & .0319 & .8454 & .8607 & .8773 & .9302 & .9319 & .9419\\
\bottomrule
\end{tabular*}
\end{minipage}\end{table*}

\section{Extended binary foreground evaluation}
\label{sec:supp-binary}

Table~\ref{tab:supp-segdec-sod} reports the complete SegDis model under
SOD-specific training. The image counts identify the evaluated image/mask
pairs. These binary metrics evaluate foreground quality only; they do not
distinguish SOD from COD and therefore complement, rather than replace, the
three-class FG-mIoU and CSCS evaluation on USC12K.

The model reaches $S_\alpha=.9398$ on ECSSD and $S_\alpha=.9224$ on DUTS-TE,
while the corresponding weighted $F_\beta$ values are .9384 and .8972.
The same rows also report maximum-threshold $F_\beta$ and $E_\phi$, so that
the table can be compared with methods that emphasize either calibrated
probabilities or thresholded masks. No per-image normalization is applied to
the probabilities before these metrics are computed.

\begin{table}[H]
\centering\footnotesize
\captionof{table}{Complete SegDis under SOD-specific training with PVT-v2-B2.
All metrics are fractions. Image counts identify the evaluated image/mask pairs.}
\label{tab:supp-segdec-sod}
\setlength{\tabcolsep}{2pt}\renewcommand{\arraystretch}{1.12}
\begin{tabular*}{\linewidth}{@{\extracolsep{\fill}}lc*{5}{c}@{}}
\toprule
\tblhead{Set} & \tblhead{\#Img} & \tblhead{$S_\alpha\uparrow$} &
\tblhead{$F_\beta^w\uparrow$} & \tblhead{MAE$\downarrow$} &
\tblhead{$F_\beta^{\max}\uparrow$} & \tblhead{$E_\phi^{\max}\uparrow$}\\
\midrule
DUTS-TE   & 5,019 & .9224 & .8972 & .0231 & .9198 & .9598\\
ECSSD     & 1,000 & .9398 & .9384 & .0240 & .9534 & .9678\\
HKU-IS    & 4,447 & .9355 & .9300 & .0207 & .9480 & .9718\\
DUT-OMRON & 5,168 & .8696 & .8088 & .0424 & .8336 & .9088\\
PASCAL-S  & 850   & .8837 & .8575 & .0493 & .8851 & .9287\\
\bottomrule
\end{tabular*}
\end{table}

\section{GT-foreground outcomes, counts, and CSCS}
\label{sec:supp-outcomes}

Table~\ref{tab:supp-outcomes} gives the breakdown behind Figure~8, including
Scene~C, which the main figure omits. Each row partitions the ground-truth
foreground of one scene into pixels typed correctly, pixels typed as the other
class (swap), and pixels predicted as background (miss). Scene~C is reported
separately for its salient support, its camouflaged support, and both together.
Table~\ref{tab:supp-outcomes-counts} gives the absolute pixel counts behind
these percentages for all four scenes. Scene~D is target-free, so it has no
ground-truth foreground; its rows measure how much background is accepted as
SOD or COD.

The reduction quoted in the main paper compares Direct-3C with SegDis on
Scene~A: SOD$\to$COD swaps fall from 760{,}453 to 563{,}455 ($-25.9\%$), while
background$\to$COD errors fall from 211{,}468 to 54{,}513 ($-74.2\%$).

\begin{table*}[t]
\noindent\begin{minipage}{\textwidth}
\captionof{table}{GT-foreground outcomes (\% of ground-truth foreground) and
CSCS on the three USC12K scenes, for the direct three-class head (Direct-3C) and
SegDis. Bold marks the best value in each column, not a comparison between
scenes. CSCS is unbolded because, as discussed in the main paper, it can move
opposite to the swap and miss rates.}
\label{tab:supp-outcomes}
\centering\footnotesize\rmfamily
\setlength{\tabcolsep}{3pt}\renewcommand{\arraystretch}{1.10}
\begin{tabular*}{\textwidth}{@{\extracolsep{\fill}}llcccccccc@{}}
\toprule
& & \multicolumn{4}{c}{Direct-3C} & \multicolumn{4}{c}{SegDis} \\
\cmidrule(lr){3-6}\cmidrule(l){7-10}
\tblhead{Scene} & \tblhead{GT support} & \tblhead{Correct$\uparrow$} & \tblhead{Swap$\downarrow$} & \tblhead{Miss$\downarrow$} & \tblhead{CSCS}
& \tblhead{Correct$\uparrow$} & \tblhead{Swap$\downarrow$} & \tblhead{Miss$\downarrow$} & \tblhead{CSCS} \\
\midrule
A & SOD & \textbf{87.38} & 4.08 & \textbf{8.54} & 39.12 & \textbf{89.66} & \textbf{3.03} & \textbf{7.31} & 45.59 \\
B & COD & 73.09 & \textbf{2.82} & 24.09 & 21.98 & 77.48 & 3.27 & 19.24 & 18.58 \\
\addlinespace[2pt]
C & SOD & 77.27 & 6.42 & 16.31 & 13.92 & 79.01 & 4.65 & 16.34 & 11.82 \\
C & COD & 62.70 & 18.35 & 18.95 & 13.92 & 63.57 & 19.02 & 17.41 & 11.82 \\
C & All FG & 73.77 & 9.29 & 16.94 & 13.92 & 75.30 & 8.10 & 16.60 & 11.82 \\
\bottomrule
\end{tabular*}
\end{minipage}
\end{table*}

\begin{table}[!t]
\captionof{table}{Absolute pixel counts behind Table~\ref{tab:supp-outcomes} for
SegDis on the four USC12K scenes. Swaps are ground-truth foreground pixels
predicted as the other class; misses are ground-truth foreground predicted as
background; BG false positives are background pixels predicted as SOD or COD.
Scene~D has no ground-truth foreground, so only its last three rows are
non-zero.}
\label{tab:supp-outcomes-counts}
\centering\scriptsize\rmfamily
\setlength{\tabcolsep}{2pt}\renewcommand{\arraystretch}{1.12}
\begin{tabular*}{\columnwidth}{@{\extracolsep{\fill}}lrrrr@{}}
\toprule
\tblhead{Quantity (px)} & \tblhead{A} & \tblhead{B} & \tblhead{C} & \tblhead{D} \\
\midrule
GT foreground & 18{,}623{,}181 & 12{,}973{,}040 & 18{,}362{,}574 & 0 \\
\quad GT SOD & 18{,}623{,}181 & 0 & 13{,}954{,}156 & 0 \\
\quad GT COD & 0 & 12{,}973{,}040 & 4{,}408{,}418 & 0 \\
GT background & 92{,}890{,}419 & 98{,}540{,}560 & 93{,}151{,}026 & 111{,}513{,}600 \\
\midrule
Correct, SOD & 16{,}698{,}200 & 0 & 11{,}025{,}215 & 0 \\
Correct, COD & 0 & 10{,}051{,}974 & 2{,}802{,}430 & 0 \\
Swaps, total & 563{,}455 & 424{,}776 & 1{,}487{,}547 & 0 \\
\quad SOD$\to$COD & 563{,}455 & 0 & 649{,}233 & 0 \\
\quad COD$\to$SOD & 0 & 424{,}776 & 838{,}314 & 0 \\
Misses, total & 1{,}361{,}526 & 2{,}496{,}290 & 3{,}047{,}382 & 0 \\
\quad SOD$\to$BG & 1{,}361{,}526 & 0 & 2{,}279{,}708 & 0 \\
\quad COD$\to$BG & 0 & 2{,}496{,}290 & 767{,}674 & 0 \\
BG false positives & 1{,}516{,}104 & 2{,}148{,}891 & 988{,}731 & 184{,}234 \\
\quad BG$\to$SOD & 1{,}461{,}591 & 718{,}412 & 603{,}388 & 16{,}939 \\
\quad BG$\to$COD & 54{,}513 & 1{,}430{,}479 & 385{,}343 & 167{,}295 \\
\bottomrule
\end{tabular*}
\end{table}

\section{Conditional type responses and label reversals}
\label{sec:supp-reversals}

The main paper analyzes type reversals using the decomposition
$p_{\mathrm{SOD}}=p_u\pi_{\mathrm{SOD}}$ and
$p_{\mathrm{COD}}=p_u\pi_{\mathrm{COD}}$. The figures in this section show
the individual terms behind that composition. Each row contains the input,
the ground-truth type map, the unified foreground support, the two conditional
type responses, and the final three-class output. Red denotes SOD, green COD,
and black background. The support map can therefore be correct even when the
conditional type preference is reversed.

\begin{figure}[!t]
\centering
\begin{subfigure}{\linewidth}
\centering
\includegraphics[width=\linewidth]{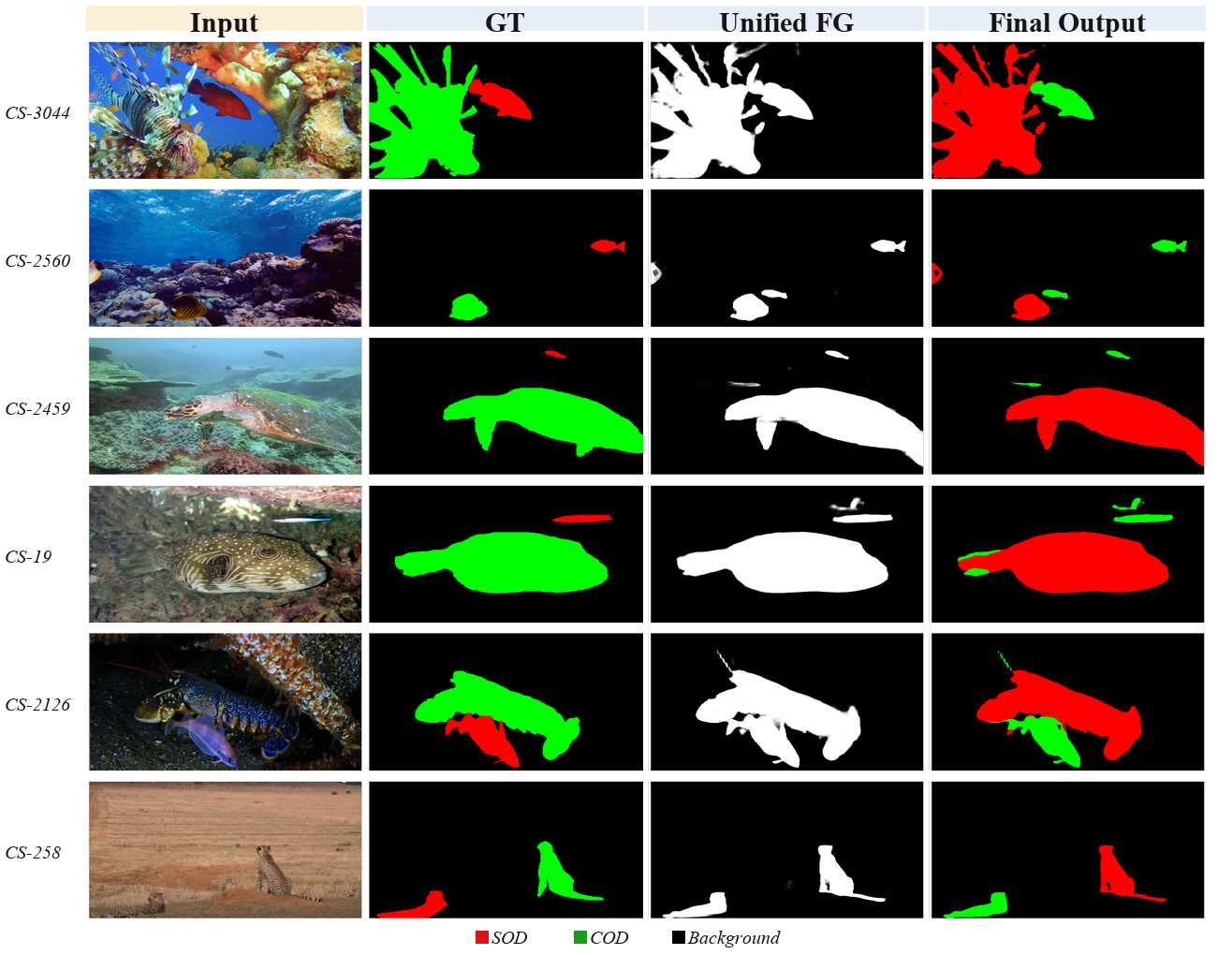}
\caption{Cases CS-3044, CS-2560, CS-2459, CS-19, CS-2126 and CS-258.}
\label{fig:supp-type-prob-a}
\end{subfigure}

\vspace{6pt}

\begin{subfigure}{\linewidth}
\centering
\includegraphics[width=\linewidth]{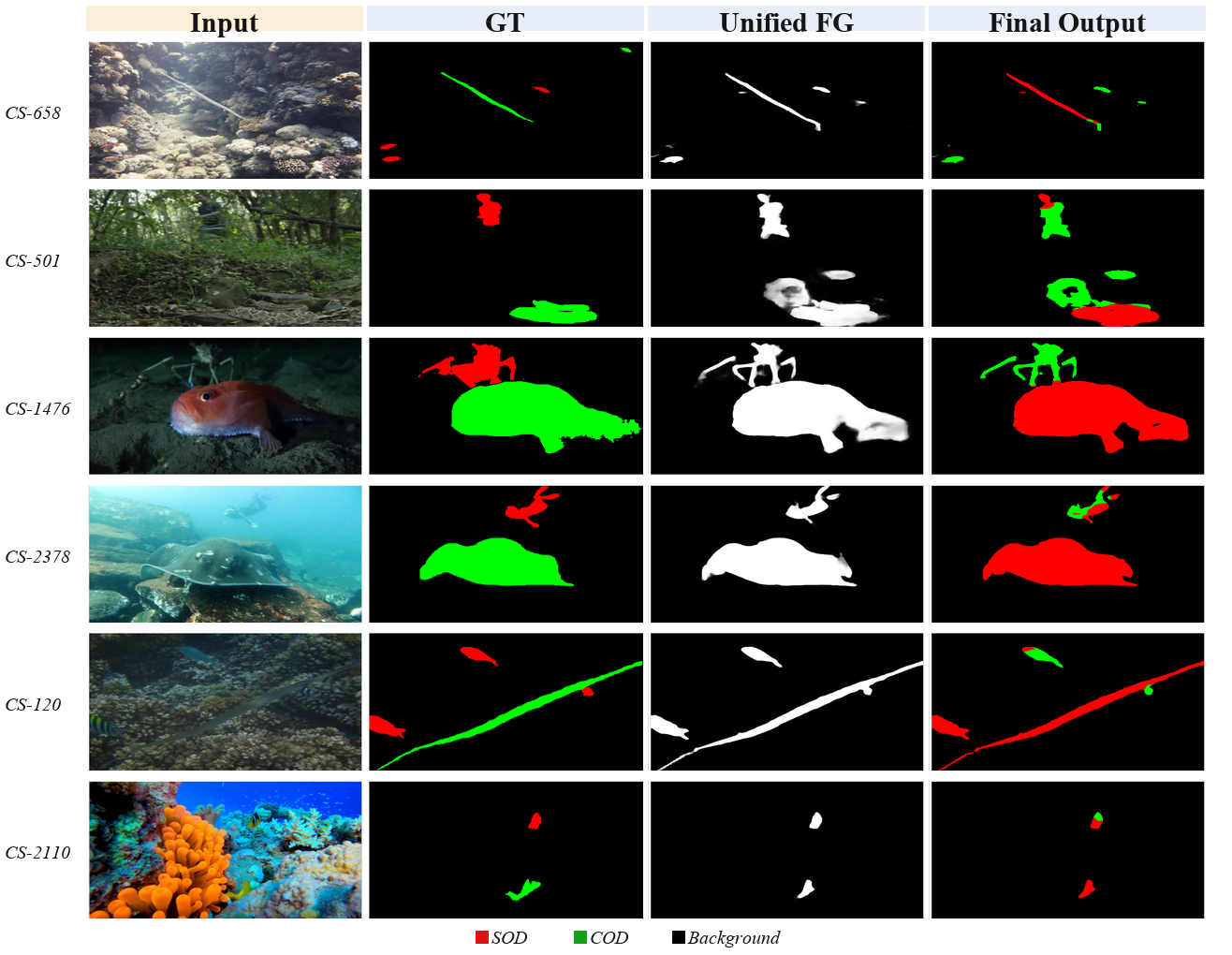}
\caption{Cases CS-658, CS-501, CS-1476, CS-2378, CS-120 and CS-2110.}
\label{fig:supp-type-prob-b}
\end{subfigure}
\caption{Conditional type responses behind label reversals. Each row is one
Scene-C case, showing the input, the ground-truth type map, the unified
foreground support, and the final output; twelve cases are shown. The support is
largely correct, while the SOD/COD allocation is reversed on the ground-truth
foreground.}
\label{fig:supp-type-prob}
\end{figure}

The examples are qualitative diagnostics rather than a prevalence estimate.
They show three distinct situations: a well-recovered support with a reversed
type, a partial type reversal together with a false positive or miss, and a
scene in which the benchmark type is difficult to infer from appearance alone.
The last case should not be interpreted as evidence of annotation error; it is
a reminder that CSCS measures agreement with benchmark labels, while the
foreground-outcome analysis measures support recovery and type assignment
separately.

\section{Master decoder and coarse prior}
\label{sec:supp-master}

The Master Decoder projects encoder features to 128 channels and fuses them
top-down,
\begin{equation}
\begin{aligned}
D_4&=\operatorname{Proj}_4(S_4),\\
D_i&=\operatorname{Fuse}_i([\operatorname{Up}(D_{i+1}),\operatorname{Proj}_i(S_i)]),\quad i=3,2,1,
\end{aligned}
\label{eq:supp-master}
\end{equation}
where $[\cdot]$ concatenates channels, projections are $1\times1$ convolutions,
and fusion is $3\times3$ Conv--GroupNorm--SiLU~\citep{wu2018groupnorm}. A head on
$D_1$ produces $z_M$ and $p_M=\sigma(z_M)$, and the coarse prior supplies
uncertainty and a blurred foreground guide,
\begin{equation}
U_M=4p_M(1-p_M),\qquad P_M=\operatorname{Blur}(p_M^{\gamma}),
\label{eq:supp-prior}
\end{equation}
with $\gamma=2$ and reflect-padded average filters of size $7\times7$.

\section{Seed variance on USC12K}
\label{sec:supp-seeds}

The model in the main paper uses seed~42. To check that the reported numbers are
not an artefact of that draw, we retrained the identical configuration with two
further seeds and evaluated each checkpoint with the same protocol; the seeds are
42, 2026, and 21. Table~\ref{tab:supp-seeds}
reports the four headline metrics per run.

\par\smallskip\noindent\begin{minipage}{\columnwidth}
\centering\scriptsize\rmfamily
\captionof{table}{Seed variance on USC12K. Each run uses the identical
architecture, data, and schedule; only the random seed differs. The last row
gives the mean and the sample standard deviation ($n-1$) over the three seeds.
Best per column in bold, with the lowest CSCS also in bold.}
\label{tab:supp-seeds}
\setlength{\tabcolsep}{1.8pt}\renewcommand{\arraystretch}{1.14}
\begin{tabular*}{\columnwidth}{@{\extracolsep{\fill}}lcccccc@{}}
\toprule
\tblhead{Seed} & \tblhead{IoU-S$\uparrow$} & \tblhead{IoU-C$\uparrow$}
& \tblhead{FG-mIoU$\uparrow$} & \tblhead{mIoU$\uparrow$} & \tblhead{mAcc$\uparrow$}
& \tblhead{CSCS$\downarrow$} \\
\midrule
42 (paper) & \textbf{75.663} & \textbf{62.304} & \textbf{68.983} & \textbf{78.351} & \textbf{85.945} & 5.752 \\
2026       & 75.651 & 61.686 & 68.668 & 78.134 & 85.407 & \textbf{5.373} \\
21         & 74.928 & 62.209 & 68.568 & 78.061 & 85.471 & 5.733 \\
\midrule
Mean & 75.414 & 62.066 & 68.740 & 78.182 & 85.608 & 5.619 \\
Std  & 0.421 & 0.333 & 0.217 & 0.151 & 0.294 & 0.213 \\
\bottomrule
\end{tabular*}
\end{minipage}\par\smallskip

The spread is small: FG-mIoU varies by 0.22 points and CSCS by 0.21 across
seeds. The gap to USCNet is 0.53 FG-mIoU, about $2.4\times$ the FG spread, and
1.74 CSCS, about eight times the CSCS spread. We report the seed-42 run in the
main paper and do not average, since all three runs support the same
conclusion.

\section{Paired comparison with the USCNet repository model}

\begin{figure*}[!t]
\centering
\includegraphics[width=0.86\textwidth]{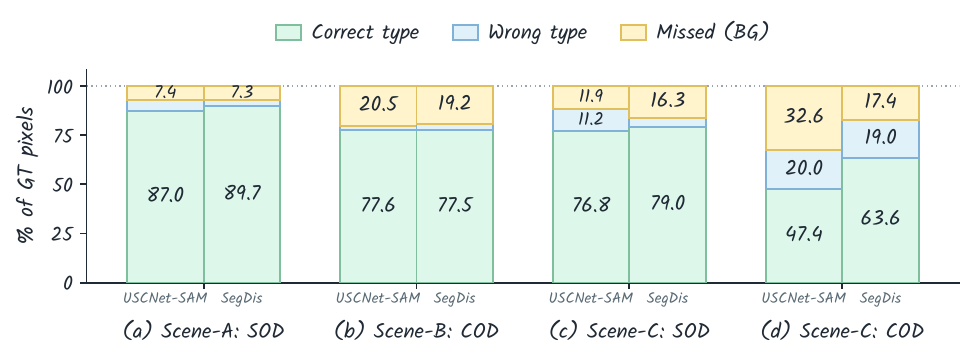}
\caption{Where ground-truth foreground pixels go, in \% of the ground-truth
pixels of that class. A correct and a swapped prediction both count as
foreground in the binary metrics; only the swapped ones enter CSCS.}
\label{fig:supp-outcome-decomp}
\end{figure*}

\label{sec:supp-uscnet-rep}

The USCNet entries in the main paper come from the ICCV~2025 paper, whose model
uses a \textbf{SAM2 Hiera-L} backbone and reports \textbf{4.04M trained
parameters} without a full-model total. The project's public repository instead
ships a \textbf{SAM ViT-H} build (640.07M parameters for the complete model),
which is a different configuration and cannot stand in for the reported one. We
therefore re-ran that repository checkpoint and our model on the same 3,600
validation images, at $352\times352$, under the three-class protocol of the main
paper. Table~S11 reports the result; the two USCNet rows
are kept apart because they are different models.

\par\smallskip\noindent{\scriptsize\rmfamily
\refstepcounter{table}\label{tab:supp-uscnet-rep}%
\textbf{Table S11.}~USCNet and SegDis under one protocol.
\emph{USCNet (Hiera-L)}: the authors' configuration, 4.04M trained parameters.
\emph{USCNet-SAM}: the repository's \textbf{SAM ViT-H} build (640.07M), re-run by
us. Best value in bold.\par}
\vspace{2pt}
\noindent{\scriptsize\rmfamily
\setlength{\tabcolsep}{0.9pt}\renewcommand{\arraystretch}{1.14}
\begin{tabular*}{\columnwidth}{@{\extracolsep{\fill}}lcccccc@{}}
\toprule
\tblhead{Model} & \tblhead{FG-mIoU$\uparrow$} & \tblhead{mIoU$\uparrow$}
& \tblhead{mAcc$\uparrow$} & \tblhead{IoU-S$\uparrow$} & \tblhead{IoU-C$\uparrow$}
& \tblhead{CSCS$\downarrow$} \\
\midrule
USCNet (Hiera-L) & 68.46 & 78.03 & 87.92 & 75.57 & 61.34 & 7.49 \\
USCNet-SAM & 65.20 & 75.85 & 83.84 & 73.86 & 56.53 & 9.82 \\
\textbf{SegDis} & \textbf{68.98} & \textbf{78.35} & \textbf{85.94}
& \textbf{75.66} & \textbf{62.30} & \textbf{5.75} \\
\bottomrule
\end{tabular*}\par}
\vspace{3pt}

Against the repository build, SegDis leads on every aggregate metric, including
mAcc (85.94 against 83.84) and CSCS (5.75 against 9.82). The paper's 87.92 mAcc
is not reproduced by the release, so the gap between the first and second rows
is a model difference rather than an evaluation difference.

\par\smallskip\noindent\begin{minipage}{\columnwidth}
\centering\scriptsize\rmfamily
\captionof{table}{Per-scene results for the two locally evaluated models. Scene~A
is SOD only, B COD only, C both types, and D background only. BG-FP is the
fraction of ground-truth background predicted as foreground.}
\label{tab:supp-uscnet-scenes}
\setlength{\tabcolsep}{2pt}\renewcommand{\arraystretch}{1.14}
\begin{tabular*}{\columnwidth}{@{\extracolsep{\fill}}llcc@{}}
\toprule
\tblhead{Scene} & \tblhead{Metric} & \tblhead{USCNet-SAM} & \tblhead{SegDis} \\
\midrule
A & IoU-S$\uparrow$ & 80.59 & \textbf{83.14} \\
B & IoU-C$\uparrow$ & \textbf{71.79} & 69.79 \\
C & IoU-S$\uparrow$ & 69.76 & \textbf{71.61} \\
C & IoU-C$\uparrow$ & 33.57 & \textbf{51.49} \\
C & FG-mIoU$\uparrow$ & 51.66 & \textbf{61.55} \\
C & CSCS$\downarrow$ & 23.62 & \textbf{11.82} \\
D & BG-FP$\downarrow$ & \textbf{0.106} & 0.165 \\
\midrule
All & binary FG IoU$\uparrow$ & \textbf{78.94} & 78.57 \\
All & FG recall$\uparrow$ & 85.70 & \textbf{86.18} \\
All & BG-FP$\downarrow$ & \textbf{1.08} & 1.22 \\
C & binary FG IoU$\uparrow$ & \textbf{79.65} & 79.14 \\
\bottomrule
\end{tabular*}
\end{minipage}\par\smallskip

The two models localise foreground almost equally well, and USCNet's binary IoU
and background false-positive rate are slightly better. The advantage of SegDis
is concentrated in Scene~C, where the two types coexist: IoU-C rises from 33.57
to 51.49 and CSCS halves. Binary results are obtained by merging the final
three-class prediction, not by thresholding $p_0$ or $p_u$.

\par\smallskip\noindent\begin{minipage}{\columnwidth}
\centering\scriptsize\rmfamily
\captionof{table}{Where the ground-truth foreground pixels go, in \% of the
ground-truth pixels of that class. Correct, swapped (predicted as the other
type), and missed (predicted as background) sum to 100 per row.}
\label{tab:supp-uscnet-outcomes}
\setlength{\tabcolsep}{2pt}\renewcommand{\arraystretch}{1.14}
\begin{tabular*}{\columnwidth}{@{\extracolsep{\fill}}llccc@{}}
\toprule
\tblhead{Scene / class} & \tblhead{Model} & \tblhead{Correct$\uparrow$}
& \tblhead{Swapped$\downarrow$} & \tblhead{Missed$\downarrow$} \\
\midrule
A / SOD & USCNet-SAM & 87.03 & 5.58 & 7.39 \\
A / SOD & SegDis & \textbf{89.66} & \textbf{3.03} & \textbf{7.31} \\
B / COD & USCNet-SAM & \textbf{77.58} & \textbf{1.89} & 20.52 \\
B / COD & SegDis & 77.48 & 3.27 & \textbf{19.24} \\
C / SOD & USCNet-SAM & 76.85 & 11.20 & \textbf{11.95} \\
C / SOD & SegDis & \textbf{79.01} & \textbf{4.65} & 16.34 \\
C / COD & USCNet-SAM & 47.40 & 19.99 & 32.60 \\
C / COD & SegDis & \textbf{63.57} & \textbf{19.02} & \textbf{17.41} \\
\bottomrule
\end{tabular*}
\end{minipage}\par\smallskip

Two effects deserve separate statements. In Scene~C the SOD type-swap rate falls
from 11.20 to 4.65, but the SOD miss rate rises from 11.95 to 16.34, so the type
gain is partly paid for with missed foreground. For Scene~C COD the dominant
change is the miss rate, from 32.60 to 17.41, while the swap rate moves only
from 19.99 to 19.02; the COD improvement is therefore mostly recovered
foreground rather than better type assignment.
Figure~\ref{fig:supp-outcome-decomp} shows the same decomposition
graphically: the green blocks move only slightly for Scene~C COD, while the red
blocks shrink by about 15 points.

\par\medskip
\begingroup
\footnotesize
\setlength{\bibsep}{0pt}
\bibliography{cvpr_references}
\bibliographystyle{ieeenat_fullname}
\endgroup

\else
{
\maketitle
\vspace{-12pt}
\begingroup\renewcommand\thefootnote{}\footnotetext{\scriptsize
$^{\ast}$Equal contribution. \quad $^{\dagger}$Corresponding author.}\endgroup
}

\begin{abstract}
Salient and camouflaged object detection (SOD and COD) both recover foreground
support, yet most joint methods output only a binary mask and therefore cannot
distinguish the type. USC12K distinguishes the two types, but its baseline USCNet
binds foreground existence to the type within one normalization, so the failure
source cannot be attributed. Our cross-task study
shows foreground localization transfers between SOD and COD, but type
discrimination does not: a COD-only model recovers foreground on SOD images but
its type judgement follows the training distribution, not the
foreground--background relation, so the two decisions need separate
mechanisms. SegDis factorizes the output into a foreground probability
and a conditional type probability: FLR provides task-agnostic support, and RTA refines the SOD/COD assignment from the
foreground--background relation without a task identifier. Because support is
explicit, each ground-truth foreground pixel is correct-type, type-swapped or
missed, so recovery and assignment errors are diagnosed separately. With PVT-v2-B2, SegDis attains 68.98\% FG-mIoU and
78.35\% mIoU on USC12K at one-eighth of USCNet's parameters; CSCS is 5.75.\ifdefined\reviewversion\else\ Code: \url{https://github.com/LUODN718/Seg2Dis}.\fi

\end{abstract}
\section{Introduction}
Salient Object Detection (SOD) and Camouflaged Object Detection (COD) are two
pixel-level foreground segmentation tasks central to visual perception and
scene understanding, with uses in wildlife monitoring, lesion and polyp
screening, and industrial defect inspection. Such deployments run under tight
compute budgets, so one small model covering both types is preferable to two
specialised ones. SOD aims to locate targets that attract human visual
attention, which typically differ from their surroundings in color, brightness,
or texture; COD focuses on targets concealed within the environment, whose
color, texture, and local contrast closely resemble the background, leading to
ambiguous boundaries and difficult foreground
recovery~\citep{wei2020f3net,fan2020camouflaged,fan2022concealed}. Despite these
different appearance conditions, both tasks share a fundamental structure:
recovering \textbf{foreground support} from a complex scene by determining which
pixels belong to the target foreground. Early research addressed the two tasks
independently, developing specialized models through multi-scale feature fusion,
context modeling, search-and-localization, boundary refinement, and uncertainty
estimation~\citep{wu2019cpd,zhao2020gatenet,liu2021vst,fan2020camouflaged,mei2021pfnet,pang2022zoom}.
These task-specific experts achieved strong performance within their own
appearance regimes, but the two research lines evolved along parallel tracks,
leaving transferable foreground evidence largely unexploited and making it
difficult to handle scenes in which salient and camouflaged objects coexist.

\begin{figure}[!tb]
\centering
\includegraphics[width=\linewidth]{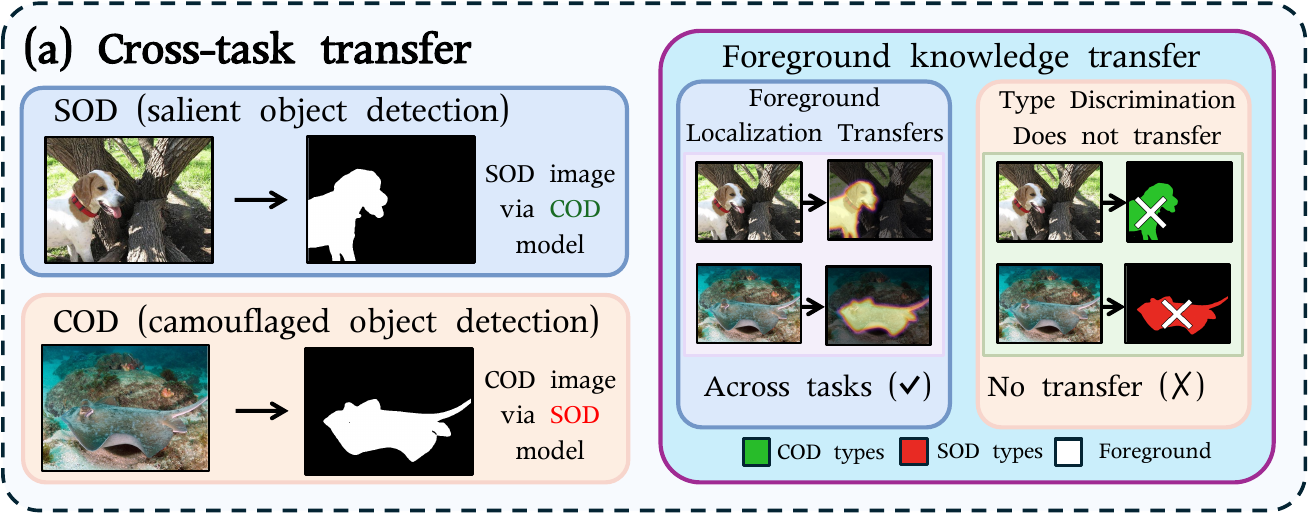}
\caption{Cross-task foreground localization and the output-space limitation. An
SOD image processed by a COD-only model and a COD image processed by an SOD-only
model both produce foreground responses. Each single-task output space contains
only its training type, so these examples illustrate the distinction between
localization transfer and explicit SOD/COD type allocation; they do not test
type transfer itself.}
\label{fig:output-space}
\end{figure}

The intuition that SOD and COD share foreground localization evidence has
motivated a line of joint learning methods. UJSC treats selected COD samples as
hard foreground instances for SOD, modeling inter-task relationships through
similarity, uncertainty, and adversarial objectives~\citep{li2021ujsc}; SENet and
JoNet further adopt shared Transformer backbones, auxiliary training, and
task-specific distribution parameters to handle both tasks within a unified
architecture~\citep{hao2025senet,hao2025distribution}. These works show that
foreground evidence can be shared across tasks, but their outputs remain binary
foreground masks. They answer \textit{where} a target is without explicitly
modeling \textit{whether} a detected foreground region is salient or
camouflaged. We call these \textbf{Untyped Joint Models}, and those that also
label the type \textbf{Typed Joint Models}. USC12K~\citep{zhou2025usc12k}
defines the typed setting over single-type, mixed, and target-free scenes, and
introduces CSCS to score SOD/COD confusion. Progress on it is still limited: the
strongest reported typed baseline reaches a CSCS of 7.49 with $228.44$M
parameters on a SAM2 Hiera-L backbone, leaving room for a smaller unified model.

To examine this relationship, we train a COD-only model and evaluate it
zero-shot on five SOD test sets, partitioned into animal and non-animal subsets.
The model approaches SOD-specific performance on animal scenes but falls short
on non-animal images (Figure~\ref{fig:category-transfer}). This indicates task-agnostic animal
foreground features, a prior that favours unified modelling. Yet the model simultaneously labels salient objects in SOD images as
camouflaged (Figure~\ref{fig:output-space}), which violates the deployment
definition of a COD model: it is deployed to identify camouflaged targets, not
to assign all foreground to the camouflaged class. The implication is that
shared foreground localization can serve as a prior for unification, but type
discrimination does not emerge from it naturally and must be introduced as an
explicit design objective.

The root cause is that salience and camouflage are relations between an object
and its surroundings, so the type cannot be read off the segmented foreground
shape. Figure~\ref{fig:output-space} shows the consequence: a model
trained only on COD data correctly localizes the foreground of an SOD image but
labels it as camouflaged,
and the resulting mask is identical to a correct detection, so the type error
never enters any binary score. Type discrimination must therefore follow the
foreground decision, and it must be produced independently, conditioned on the
foreground support.

We thus propose SegDis, a unified predictor that keeps the two answers apart.
Foreground Localization Refinement (FLR) recovers a high-quality,
task-agnostic \textbf{foreground support} map through coarse-to-fine residual
refinement; its only role is to decide whether a pixel belongs to the
foreground. Relation-conditioned Type Allocation (RTA) then uses that support to
gate foreground--background relational pooling, dynamically modulates SOD/COD
type queries, and assigns a per-pixel \textbf{conditional type} without
requiring a task identifier at inference. The two probabilities compose into normalized background/SOD/COD
predictions and support mixed scenes, and because the foreground probability is
output explicitly, every ground-truth foreground pixel is classified as
correct-type, type-swapped, or missed, so foreground-recovery and
type-assignment errors are diagnosed separately.

\begin{figure}[!t]
\centering
\includegraphics[width=\linewidth]{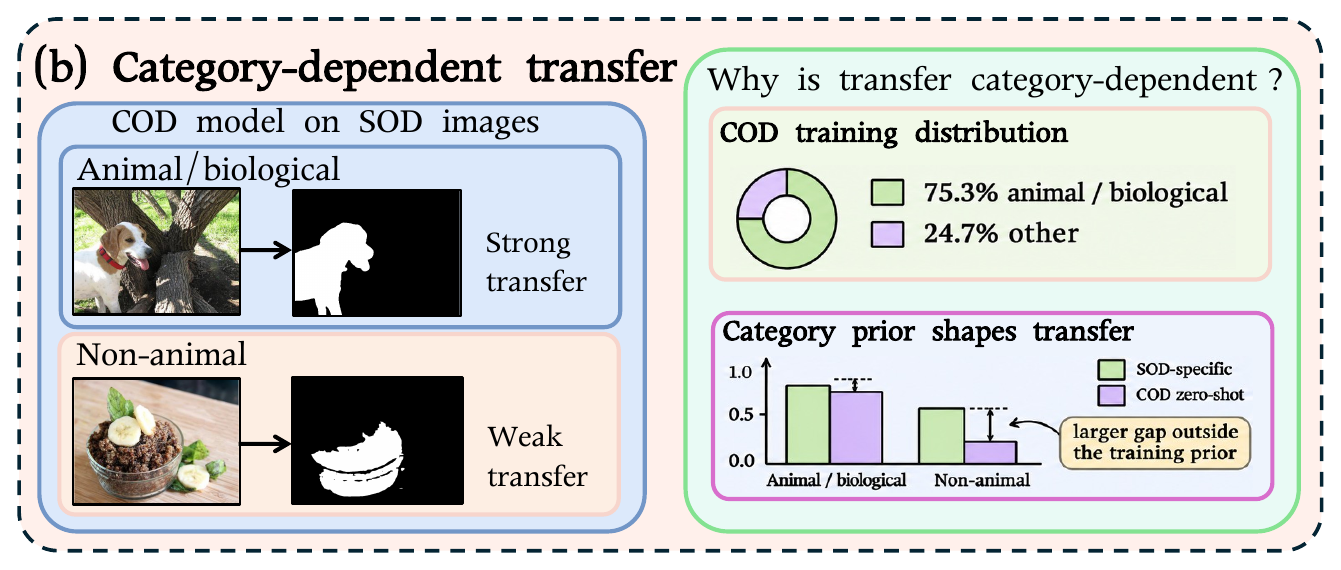}
\caption{Category-dependent transfer. A COD-only model applied to SOD images
recovers a stronger foreground response on animal scenes than on non-animal
scenes. The right panel reports the animal-heavy COD training distribution and
shows that the gap between SOD-specific performance and COD zero-shot transfer
widens outside that category mix.}
\label{fig:zero-shot-transfer}
\end{figure}

\begin{figure}[!t]
\centering
\includegraphics[width=\linewidth]{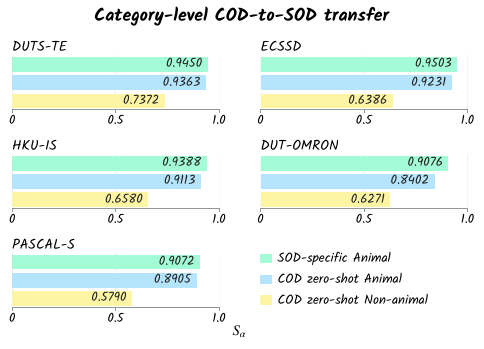}
\caption{Category-level COD-to-SOD transfer. Animal subsets approach
SOD-specific performance, while non-animal performance is lower.}
\label{fig:category-transfer}
\end{figure}

\noindent
\textbf{Contributions.}
\begin{itemize}\setlength{\itemsep}{2pt}
\item We show in a COD-to-SOD transfer study on five SOD benchmarks that foreground localization is category-dependent, while a single-task binary output cannot resolve the foreground type without an explicit type-allocation mechanism (\S\ref{sec:experiments}).
\item We propose SegDis, which separates foreground support estimation (FLR) from conditional type assignment (RTA) and assigns per-pixel conditional types without a task identifier at inference, so that mixed scenes are parsed as a support decision followed by a type decision.
\item Relative to a direct three-class head, our factorized path reduces type confusion while maintaining competitive binary foreground segmentation; the resulting model uses a 28.80M-parameter PVT-v2-B2 encoder.
\end{itemize}

\section{Related Work}
\subsection{Task-Specific Models}
SOD and COD were developed as separate model families. SOD moved from contrast
cues to multiscale fusion and detail discovery, with GateNet and ICON improving
information flow and VST and PGNet broadening contextual
modeling~\citep{yan2013ecssd,yang2013dutomron,wu2019cpd,wei2020f3net,zhao2020gatenet,zhuge2022icon,liu2021vst,xie2022pgnet}.
COD targets weak foreground--background differences, from SINet's
search-and-identification reasoning to distraction mining, mixed-scale
processing, Transformer backbones, and frequency-aware
designs~\citep{fan2020camouflaged,fan2022concealed,mei2021pfnet,pang2022zoom,huang2023fspnet,yin2024camoformer,he2023feder,cong2023fpnet,sun2024fsel}.
Each family is tuned to one appearance regime and is not applied to the other,
even though both recover a shared foreground support.

\subsection{Untyped Joint Models}
A second line shares foreground evidence but stops at a binary mask. UJSC turns
the task contradiction into a training signal, using selected COD samples as
hard SOD instances with similarity, uncertainty, and adversarial
objectives~\citep{li2021ujsc}; SENet and JoNet share Transformer backbones with
auxiliary training, task-specific distribution parameters, and saliency-aware
sampling~\citep{hao2025senet,hao2025distribution}; Seamless Detection relies on
contrastive distillation and contextual decoding~\citep{liu2025seamless}.
Because the output is one mask, these models answer \textit{where} but not \textit{whether}.

\subsection{Typed Joint Models}
A third line emits a type label together with the mask. USC12K defines
background/SOD/COD parsing over single-type, mixed, and target-free scenes and
introduces CSCS for type confusion~\citep{zhou2025usc12k}. Its baseline USCNet
conditions a frozen SAM mask decoder on three learned aspect prompts (salient,
camouflaged, background) and applies a softmax over the three masks, so
foreground existence and type share one normalized competition. SegDis keeps
this typed setting but splits the output into a support probability
$P(\mathrm{FG})$ and a conditional type distribution
$P(\mathrm{type}\mid\mathrm{FG})$, so foreground recovery and type-assignment
errors can be diagnosed apart. Our Direct-3C head is an in-house diagnostic
baseline.

\section{Method}
\label{sec:method}

\subsection{Factorized Prediction}
The design separates two uncertainties: whether a pixel belongs to any
foreground, and which type explains that foreground. This shares localization
across SOD-only, COD-only, and mixed scenes while making misses, false positives,
and SOD/COD swaps separately measurable. Probability composition keeps the
three outputs normalized.
SegDis predicts common foreground support and a conditional distribution
over its two types (Figure~\ref{fig:overview}). FLR estimates the union
of salient and camouflaged targets. RTA uses the predicted support and
semantic features to obtain $\pi_{\mathrm{SOD}}$ and $\pi_{\mathrm{COD}}$,
with $\pi_{\mathrm{SOD}}+\pi_{\mathrm{COD}}=1$, and a bounded support
correction. Corrected foreground confidence $p_u$ gives
$p_{\mathrm{BG}}=1-p_u$ and $p_t=p_u\pi_t$ for the two foreground types.
The binary objective merges SOD/COD labels; the conditional type loss is
evaluated on ground-truth foreground pixels. Pooling at inference uses
predicted support. Both components are jointly trained in one forward
pass, and inference requires no task or scene identifier.

\begin{figure*}[t]
\begin{minipage}{\textwidth}
\centering
\includegraphics[width=.98\textwidth]{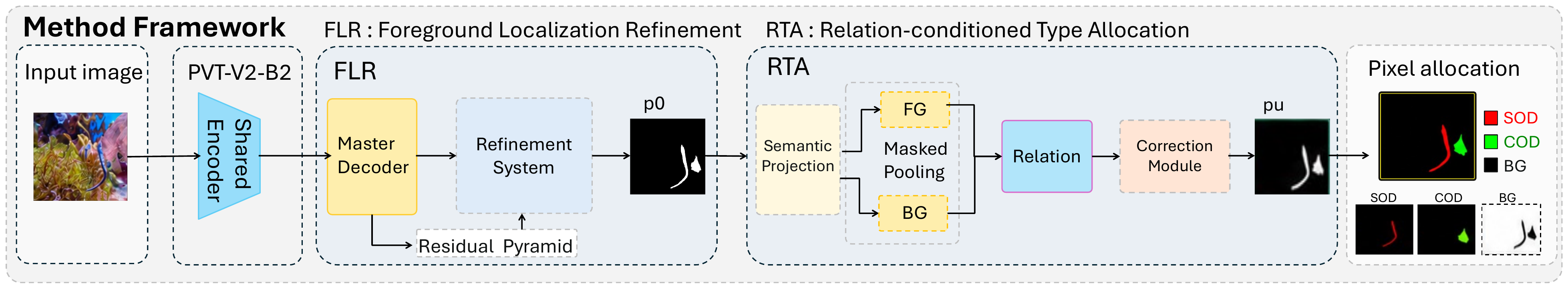}
\captionof{figure}{SegDis shares foreground localization and explicitly
conditions type discrimination on foreground--background relations.
FLR estimates common support $p_0$. RTA uses support-weighted semantic
pooling to adapt SOD/COD queries and obtain conditional type probabilities.
The corrected support $p_u$ and conditional types compose the final output.}
\label{fig:overview}
\end{minipage}
\end{figure*}

\subsection{Foreground Support}
\label{sec:flr}

\flr uses a PVT-v2-B2 encoder~\citep{wang2022pvtv2} and top-down
feature fusion~\citep{lin2017fpn}, with four features
$\{S_i\}_{i=1}^4$ of $64$, $128$, $320$, and $512$ channels. It combines a
coarse prior with residual evidence and region/boundary refinement
(Figure~\ref{fig:flr}). Within this pathway, the Master Decoder produces the
coarse prior that the residual system refines into $p_0$.

\begin{figure}[!t]
\centering
\includegraphics[width=\linewidth]{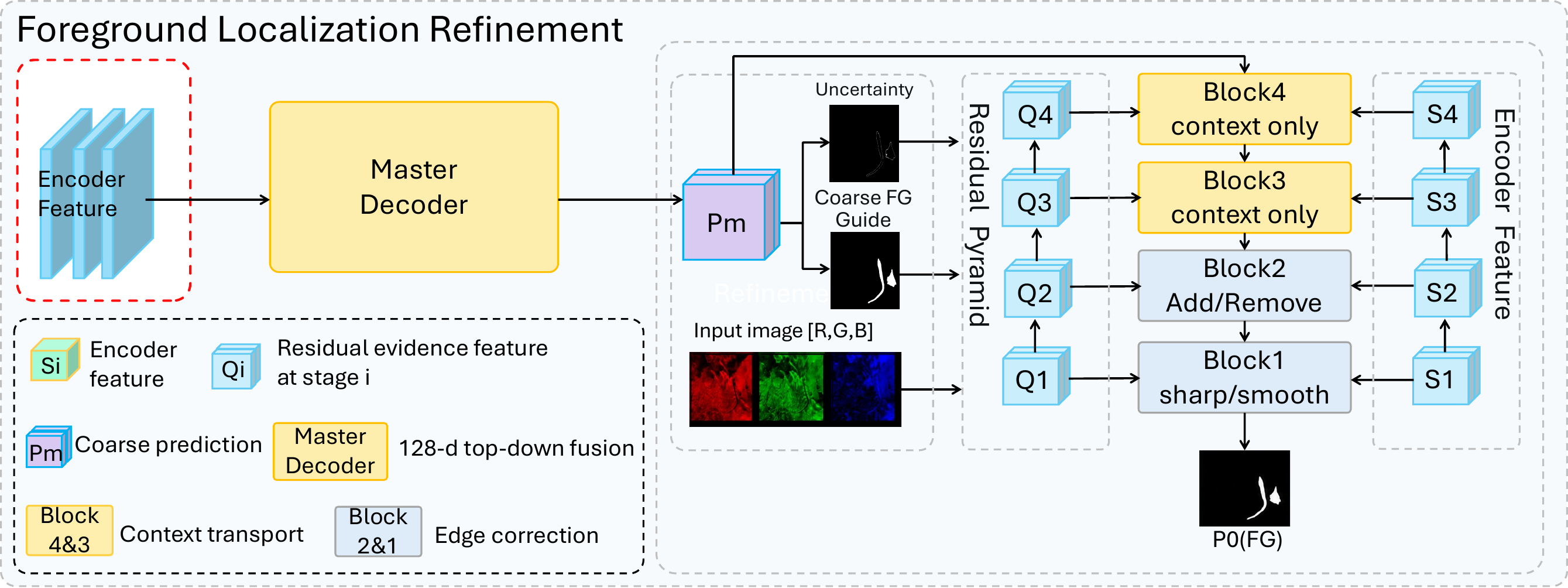}
\caption{Foreground Localization Refinement (FLR). The coarse prediction supplies a blurred foreground guide $P_M$, uncertainty $U_M$, and refinement seed $p_M$; residual features $Q_i$ and encoder skips $S_i$ support progressive context, region, and boundary refinement into $p_0$.}
\label{fig:flr}
\end{figure}

\paragraph{Coarse prior and residual evidence.}
The Master Decoder projects encoder features to 128 channels and fuses them
top-down (Supplementary Eq.~S1), with $[\cdot]$ concatenating channels, $1\times1$
projections, and $3\times3$ Conv--GroupNorm--SiLU~\citep{wu2018groupnorm} fusion.
A head on $D_1$ produces $z_M$ and $p_M=\sigma(z_M)$, from which the coarse prior
supplies uncertainty and a blurred foreground guide (Supplementary Eq.~S2).
A 48-channel residual pyramid extracts $\{Q_i\}$ at strides 4, 8, 16,
and 32 from the five-channel input $[I,P_M,U_M]$, consisting of the RGB image $I$,
a blurred coarse foreground guide $P_M$, and uncertainty $U_M$. The guide is
formed from the detached coarse probability and provides a soft foreground
prior; $U_M$ marks uncertainty. These cues are concatenated before residual
encoding.
We use $\gamma=2$ and reflect-padded average filters of size $7\times7$
for $P_M$, using detached coarse probabilities. The coupled residual
refinement pathway learns how to use these cues.

\paragraph{Progressive correction.}
Coarse stages aggregate encoder features, residual evidence, foreground
probabilities, uncertainty, and propagated context. At the region stage,
add/remove responses update the current probability:
\begin{equation}
p'=\operatorname{clip}_{[\epsilon,1-\epsilon]}\{p+\lambda[(1-p)\sigma(a)-p\sigma(d)]\}.
\end{equation}
Here $\lambda=0.25$ and $\epsilon=10^{-5}$. The add term fills missing
foreground, while the remove term suppresses excess responses.
Each pyramid block concatenates its residual evidence $Q_i$ with the matching
encoder feature $S_i$, applies a $3\times3$ Conv--GroupNorm--SiLU, and passes the
result down to $Q_{i-1}$. B4/B3 propagate coarse context and never write to the
probability map; B2 is the first block that admits the add and remove terms
above. At B1, a smooth global branch preserves spatial context, while a sharp
ROI branch uses ROI Align~\citep{he2017maskrcnn} to refine local structure. A learned
router combines their corrections to obtain $p_0$ and its logit $z_0$.
Reproducibility details are left to the released code.

Together, these stages produce a task-agnostic support map $p_0$. Its role
is deliberately limited to deciding whether a pixel belongs to foreground;
the following type module uses $p_0$ as evidence rather than as a task label.
This separation lets the same support pathway serve SOD-only, COD-only, and
mixed inputs before contextual type assignment.

\subsection{Contextual Type}
\begin{figure}[H]
\centering
\includegraphics[width=\linewidth]{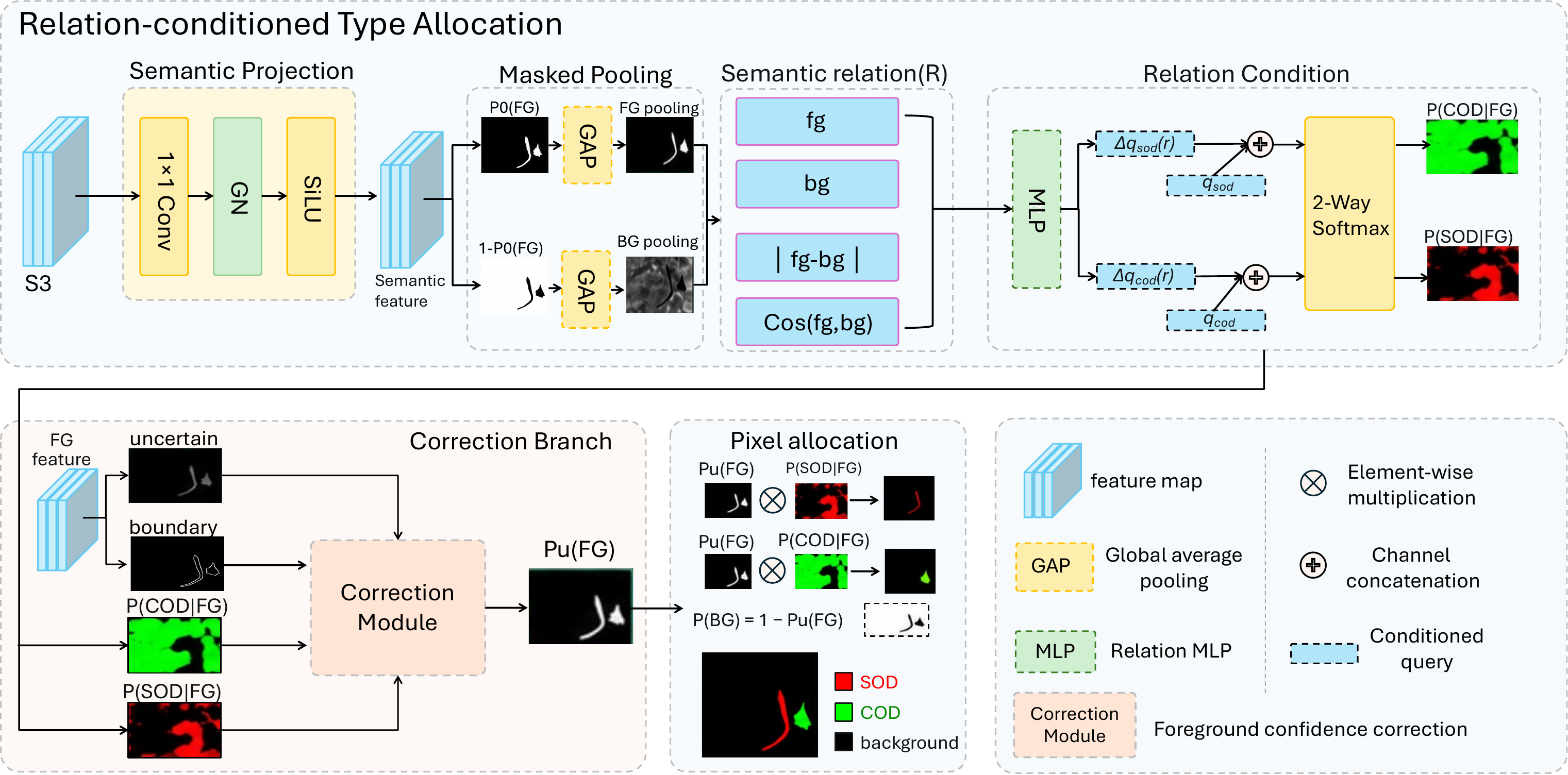}
\caption{Relation-conditioned Type Allocation (RTA) in detail. Masked pooling summarises foreground and background evidence as $f_{\mathrm{fg}}$ and $f_{\mathrm{bg}}$; semantic projection and an MLP turn them into the relation vector $r$, which conditions the SOD/COD queries and yields the conditional types $P(\mathrm{COD}\mid\mathrm{FG})$ and $P(\mathrm{SOD}\mid\mathrm{FG})$.}
\label{fig:fbrc}
\end{figure}
Within the parsing pathway, Relation-conditioned Type Allocation
(RTA) converts foreground support and semantic features into conditional
types and a bounded foreground correction. The predicted support has three
roles: it weights foreground/background semantic pooling, conditions the
type-query adaptation, and gates the final correction and probability
composition. Figure~\ref{fig:fbrc} summarizes the pathway.

\paragraph{Image-level foreground--background relation.}
A $1\times1$ Conv--GroupNorm--SiLU projects S3 to a 96-channel semantic map
$F$, resized to the resolution of $p_0$. Foreground and background soft pooling
give
\begin{equation}
\begin{aligned}
f_{\mathrm{fg}}&=\frac{\sum_x F(x)\operatorname{sg}[p_0(x)]}
{\max(\sum_x\operatorname{sg}[p_0(x)],\epsilon)},\\
f_{\mathrm{bg}}&=\frac{\sum_x F(x)\operatorname{sg}[1-p_0(x)]}
{\max(\sum_x\operatorname{sg}[1-p_0(x)],\epsilon)}.
\end{aligned}
\end{equation}
Stop-gradient $\operatorname{sg}$ fixes the pooling weights while semantic
features remain trainable. The relation
\begin{equation}
r=[f_{\mathrm{fg}},f_{\mathrm{bg}},|f_{\mathrm{fg}}-f_{\mathrm{bg}}|,
\cos(f_{\mathrm{fg}},f_{\mathrm{bg}})]\in\mathbb R^{289}
\end{equation}
summarizes image-level foreground--background semantics; local features
provide pixel-wise type responses.
The pooled relation conditions both queries on the same scene, whereas
local feature matching allows different regions to receive different
types. Thus image-level context does not impose one type on the image.

Type classification uses locally projected features $F_l(x)$:
\begin{equation}
\pi_t(x)=P(t\mid\mathrm{FG},x)=
\frac{\exp(q_t^\top F_l(x)/\tau)}
{\sum_j\exp(q_j^\top F_l(x)/\tau)}.
\end{equation}
The class weights $q_t=\bar q_t+\Delta q_t$, shown as queries in
Figure~\ref{fig:overview}, combine learned base weights with two independent
96-dimensional offsets $[\Delta q_{\mathrm{SOD}},\Delta q_{\mathrm{COD}}]=g(r)$.
Here $g$ is Linear--LayerNorm--SiLU and
$\tau=\max(|\tau_{\mathrm{learned}}|,0.25)$.
This is a linear classifier on $F_l$ with image-conditioned weights.
A common offset would cancel in the two-way softmax.

\ifdefined\rtadoublecolumn\Needspace{8\baselineskip}\fi
\paragraph{Foreground correction within RTA.}
The final correction uses local features, foreground uncertainty
$u_0=4p_0(1-p_0)$, a boundary response
$b_0=\operatorname{MaxPool}_{3\times3}(p_0)-\operatorname{MinPool}_{3\times3}(p_0)$,
and the signed type preference $\pi_{\mathrm{SOD}}-\pi_{\mathrm{COD}}$:
\begin{equation}
\begin{aligned}
\Delta z&=0.2\tanh h([F_l,u_0,b_0,\pi_{\mathrm{SOD}}-\pi_{\mathrm{COD}}]),\\
z_u&=z_0+\operatorname{sg}[(u_0+b_0)/2]\Delta z.
\end{aligned}
\label{eq:correction}
\end{equation}
The gate weights a bounded foreground update while keeping conditional
types fixed (Eq.~\ref{eq:correction}). With $p_u=\sigma(z_u)$, the final
probabilities are
\begin{equation}
p_{\mathrm{BG}}=1-p_u,\qquad
p_{\mathrm{SOD}}=p_u\pi_{\mathrm{SOD}},\qquad
p_{\mathrm{COD}}=p_u\pi_{\mathrm{COD}}.
\label{eq:composition}
\end{equation}
They sum to one (Eq.~\ref{eq:composition}), and a three-way argmax
yields the label. This is the pixel-allocation box of
Figure~\ref{fig:fbrc}: the corrected support $p_u$ scales both conditional
types, and the background probability takes up the remainder.
At a fixed forward pass, $p_{\mathrm{SOD}}/p_{\mathrm{COD}}
=\pi_{\mathrm{SOD}}/\pi_{\mathrm{COD}}$: correction changes competition
with background, while the type classifier determines the relative
SOD/COD preference. Training minimizes
\begin{equation}
\mathcal L=\mathcal L_{\mathrm{fg}}+0.5\mathcal L_{\mathrm{type}}
+0.5\mathcal L_{\mathrm{cal}}.
\end{equation}
$\mathcal L_{\mathrm{fg}}$ sums weighted BCE and weighted IoU on $p_u$,
using $w=1+5|\operatorname{AvgPool}_{31\times31}(y)-y|$ for binary target $y$.
Each term is normalized spatially per image.
$\mathcal L_{\mathrm{type}}$ averages cross-entropy over ground-truth foreground
pixels only, so the type head never models foreground existence whereas a
direct three-way head must; $\mathcal L_{\mathrm{cal}}$ averages
composed three-class negative log-likelihood over all pixels.
For a true foreground pixel
of type $t$, its contribution is $-\log p_u-\log\pi_t$, so this last term
also supervises the type branch. Common support receives binary foreground
supervision, while the type and composed losses train its SOD/COD allocation.
\FloatBarrier

\setcounter{dbltopnumber}{4}
\section{Experiments}
\label{sec:experiments}

\subsection{Experimental setup}
We evaluate unified three-class parsing on USC12K and binary foreground
segmentation under SOD-specific, COD-specific, and paired joint training.
Table~\ref{tab:experimental-setup} summarizes the training settings.

\par\smallskip
\noindent\begin{minipage}{\columnwidth}
\centering
\captionof{table}{Experimental configuration summary.}
\label{tab:experimental-setup}
\scriptsize
\renewcommand{\arraystretch}{1.15}
\setlength{\tabcolsep}{4pt}
\begin{tabular*}{\linewidth}{@{\extracolsep{\fill}}lll@{}}
\toprule
\textbf{Component} & \textbf{Configuration} & \textbf{Details} \\
\midrule
FLR training & SOD/COD, $384^2$ & Specialized or paired joint \\
SegDis training & USC12K, $352^2$ & 8,400 / 3,600 eval, end-to-end \\
\midrule
Backbone & PVT-v2-B2 & ImageNet pretrained \\
Optimizer & AdamW & lr $10^{-4}$, decay $5\times10^{-4}$ \\
Schedule & 50 epochs, batch 8 & Cosine decay, seed 42 \\
\midrule
Evaluation & 3-class argmax & Nearest-neighbor $352^2$ labels \\
\bottomrule
\end{tabular*}
\end{minipage}
\par\smallskip

\paragraph{Evaluation.}
We follow USCNet's raw three-class protocol: labels are resized by nearest
neighbor to $352\times352$, predictions use three-class argmax at this
resolution, and confusion matrices are accumulated over the evaluation split.
No Scene-A/B label remapping or scene identifier is used.
We report FG-mIoU, mIoU, mAcc, IoU-S, IoU-C, and CSCS.
Binary foreground quality is measured by $S_\alpha$, weighted
$F_\beta$, MAE, maximum $F_\beta$, and maximum $E_\phi$~\citep{fan2017smeasure,margolin2014weightedf,fan2018emeasure}.
The SOD tests use DUTS-TE, ECSSD, HKU-IS, DUT-OMRON, and
PASCAL-S~\citep{wang2017duts,yan2013ecssd,li2015hkis,yang2013dutomron,li2014pascals};
COD comparisons use CAMO, COD10K, and NC4K~\citep{le2019camo,fan2020camouflaged,lv2021nc4k},
with additional task-specific evaluation on CHAMELEON~\citep{skurowski2018chameleon}.
All parameter counts reported in this paper are total parameters, including any
frozen backbone, which is the convention used by the USC12K comparison as well as
by the cost table below.

\subsection{Unified SOD/COD parsing}
\begin{table*}[!t]
\noindent\begin{minipage}{\textwidth}
\captionof{table}{Comparison on the USC12K test set (\%). External scores and
the parameter column reproduce Table~3 of \citet{zhou2025usc12k}; unless marked,
the column denotes total model parameters including frozen backbones.
$^{\ddagger}$~full-model estimate: 224.4M for the SAM~2.1 Hiera-L
checkpoint~\citep{ravi2024sam2} plus USCNet's 4.04M trained parameters; only the
latter is reported by USCNet.
FG-mIoU averages the displayed overall IoU-S and IoU-C, rounded to two
decimals. Higher is better except CSCS; best values are bold. -- denotes an
unreported value.}\label{tab:usc-full}
\centering
\footnotesize
\setlength{\tabcolsep}{1.0pt}
\renewcommand{\arraystretch}{1.0}
\begin{tabular*}{\textwidth}{@{\extracolsep{\fill}}llc*{6}{r}*{4}{c}@{}}
\toprule

& & & \multicolumn{1}{c}{Scene A} & \multicolumn{1}{c}{Scene B} & \multicolumn{2}{c}{Scene C}
& \multicolumn{6}{c}{Overall scenes} \\
\cmidrule(lr){4-4}\cmidrule(lr){5-5}\cmidrule(lr){6-7}\cmidrule(l){8-13}

\textbf{Method} & \textbf{Backbone} & \textbf{Total params (M)} & \textbf{IoU-S} & \textbf{IoU-C} & \textbf{IoU-S} & \textbf{IoU-C} & \textbf{IoU-S} & \textbf{IoU-C} & \textbf{FG-mIoU} & \textbf{mIoU} & \textbf{mAcc} & \textbf{CSCS} \\
\midrule
\multicolumn{13}{@{}l}{\textcolor{tblaccent}{\textbf{SOD-specific methods}}} \\
\modelcite{GateNet}{zhao2020gatenet} & ResNeXt-101 & $128.63$ & 68.32 & 54.26 & 66.85 & 35.03 & 65.08 & 44.17 & 54.63 & 68.27 & 78.07 & 11.30 \\
\modelcite{F3Net}{wei2020f3net} & ResNet-50 & $25.54$ & 70.05 & 52.62 & 67.20 & 36.38 & 66.12 & 44.81 & 55.47 & 68.80 & 77.86 & 9.36 \\
\modelcite{Auto-MSFNet}{zhang2021msfnet} & ResNet-50 & $27.83$ & 70.14 & 54.78 & 69.92 & 36.64 & 66.69 & 45.89 & 56.29 & 69.40 & 79.77 & 9.90 \\
\modelcite{VST}{liu2021vst} & T2T-ViT-t-14 & $44.48$ & 68.14 & 49.82 & 61.61 & 22.56 & 63.18 & 38.45 & 50.82 & 65.55 & 74.77 & 11.30 \\
\modelcite{EDN}{wu2022edn} & ResNet-50 & $43.20$ & 71.59 & 57.94 & 69.37 & 37.70 & 68.00 & 48.27 & 58.14 & 70.70 & 80.60 & 9.23 \\
\modelcite{PGNet}{xie2022pgnet} & R18 + Swin-B & $72.67$ & 74.69 & 57.31 & 71.94 & 37.21 & 70.72 & 48.78 & 59.75 & 71.82 & 80.76 & 7.71 \\
\modelcite{ICON-R}{zhuge2022icon} & ResNet-50 & $33.09$ & 68.09 & 50.57 & 67.48 & 30.65 & 65.86 & 45.53 & 55.70 & 68.99 & 79.53 & 10.24 \\
\modelcite{MDSAM}{gao2024mdsam} & SAM ViT-B & $100.21$ & 72.96 & 56.16 & 67.21 & 36.06 & 69.67 & 49.05 & 59.36 & 71.67 & 82.92 & 10.21 \\
\midrule
\multicolumn{13}{@{}l}{\textcolor{tblaccent}{\textbf{COD-specific methods}}} \\
\modelcite{SINet-V2}{fan2022concealed} & Res2Net-50 & $26.96$ & 72.96 & 56.16 & 67.21 & 36.06 & 69.50 & 47.47 & 58.49 & 70.20 & 79.58 & 8.83 \\
\modelcite{PFNet}{mei2021pfnet} & ResNet-50 & $46.50$ & 69.07 & 52.83 & 67.20 & 32.81 & 65.73 & 43.76 & 54.75 & 68.30 & 78.00 & 10.04 \\
\modelcite{ZoomNet}{pang2022zoom} & ResNet-50 & $33.35$ & 74.11 & 51.12 & 66.79 & 29.69 & 66.43 & 43.28 & 54.86 & 68.35 & 77.72 & 8.88 \\
\modelcite{FEDER}{he2023feder} & ResNet-50 & $44.10$ & 74.35 & 58.04 & 67.66 & 32.26 & 68.65 & 46.46 & 57.56 & 70.32 & 81.27 & 10.01 \\
\modelcite{PRNet}{hu2024prnet} & SMT-T & $12.74$ & 76.10 & 61.54 & 60.10 & 32.16 & 68.68 & 50.88 & 59.78 & 71.87 & 82.89 & 8.40 \\
\modelcite{ICEG}{he2024iceg} & Swin-B & $99.60$ & 73.67 & 68.38 & 68.43 & 44.33 & 69.22 & 58.71 & 63.97 & 74.68 & 83.53 & 8.16 \\
\modelcite{CamoDiffusion}{chen2024camodiffusion} & PVT-v2-B4 & $71.80$ & 75.01 & 59.39 & 53.49 & 45.03 & 63.49 & 52.80 & 58.15 & 70.70 & 77.73 & 7.73 \\
\modelcite{CamoFormer-P}{yin2024camoformer} & PVT-v2-B4 & $71.00$ & 75.88 & 66.19 & 73.33 & 44.14 & 71.86 & 56.09 & 63.98 & 74.81 & 84.17 & 7.57 \\
\modelcite{PGT-P4}{wang2024pgt} & PVT-v2-B4 & $68.20$ & 72.75 & 61.51 & 70.01 & 41.21 & 71.46 & 56.83 & 64.15 & 75.03 & 83.35 & 9.09 \\
\modelcite{SAM-Adapter}{chen2023samadapter} & SAM ViT-H & $641.11$ & 78.90 & 67.69 & 68.19 & 27.73 & 70.66 & 52.69 & 61.68 & 73.38 & 83.35 & 10.28 \\
\modelcite{SAM2-Adapter}{chen2024sam2adapter} & SAM2 Hiera-L & $228.76$ & 78.75 & 70.28 & 69.01 & 38.20 & 71.42 & 56.71 & 64.07 & 74.98 & 84.74 & 9.12 \\
\midrule
\multicolumn{13}{@{}l}{\textcolor{tblaccent}{\textbf{Unified SOD/COD methods}}} \\
\modelcite{EVP}{liu2023evp} & SegFormer-B4 & $69.07$ & 75.85 & 59.81 & 71.41 & 37.64 & 70.30 & 50.36 & 60.33 & 72.16 & 79.96 & 8.67 \\
\modelcite{VSCode}{luo2024vscode} & Swin-T & $54.09$ & 76.04 & 60.31 & 72.58 & 39.46 & 71.08 & 55.14 & 63.11 & 74.17 & 84.01 & 8.17 \\
\modelcite{Spider}{zhao2024spider} & ConvNeXt-B & $175.20$ & 76.81 & 63.54 & 72.87 & 42.39 & 71.34 & 56.68 & 64.01 & 74.92 & 85.79 & 7.86 \\
\modelcite{USCNet}{zhou2025usc12k} & SAM2 Hiera-L & $228.44^{\ddagger}$ & 79.70 & \textbf{74.99} & \textbf{74.80} & 45.73 & 75.57 & 61.34 & 68.46 & 78.03 & \textbf{87.92} & 7.49 \\
\midrule

\textbf{\method (Ours)} & \textbf{PVT-v2-B2} & \textbf{28.80} &
83.14 & 69.79 & 71.61 & 51.49 &
75.66 & 62.30 & 68.98 & 78.35 & 85.95 & 5.75 \\
\method (Ours) & PVT-v2-B4 & 66.00 &
\textbf{84.14} & 69.45 & 72.05 & \textbf{53.91} &
\textbf{76.22} & \textbf{63.31} & \textbf{69.77} & \textbf{78.91} & 85.71 & \textbf{5.43} \\
\bottomrule
\end{tabular*}
\end{minipage}\par
\end{table*}

\begin{figure*}[!t]
\begin{minipage}{\textwidth}
\centering
\includegraphics[width=\textwidth]{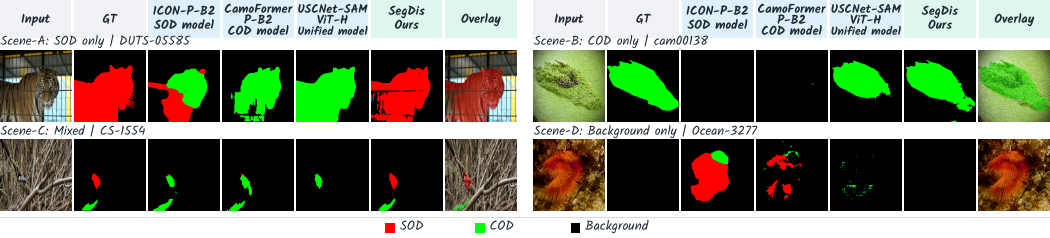}
\captionof{figure}{Qualitative comparison with all four models trained on USC12K. ICON-P-B2 and CamoFormer-P-B2 were originally designed for SOD and COD, respectively; USCNet-SAM (ViT-H) is a unified baseline, and SegDis is ours. Overlays show SegDis predictions at 42\% opacity.}
\label{fig:quality-predictions}
\end{minipage}
\end{figure*}

\paragraph{Localization and type metrics.}
From global confusion counts, $\mathrm{IoU}_c=\mathrm{TP}_c/
(\mathrm{TP}_c+\mathrm{FP}_c+\mathrm{FN}_c)$. We compute
\begin{equation}
\begin{aligned}
\mathrm{FG\text{-}mIoU}&=(\mathrm{IoU}_{\mathrm S}+\mathrm{IoU}_{\mathrm C})/2,\\
\mathrm{mIoU}&=(\mathrm{IoU}_{\mathrm{BG}}+\mathrm{IoU}_{\mathrm S}+\mathrm{IoU}_{\mathrm C})/3,\\
\mathrm{mAcc}&=\tfrac13\sum_{c\in\{\mathrm{BG,S,C}\}}
\frac{\mathrm{TP}_c}{\mathrm{TP}_c+\mathrm{FN}_c}.
\end{aligned}
\end{equation}
Binary foreground IoU merges SOD and COD, so a type-swapped mask can still
overlap well. CSCS~\citep{zhou2025usc12k} normalises each cross-type confusion by the pixels
\emph{predicted} as that class, so it behaves like precision: predicting more
foreground enlarges the denominator and can lower it without better
discrimination. We therefore also split GT foreground by outcome, with
denominators fixed by the annotation.

Table~\ref{tab:usc-full} compares SegDis with 23 methods from
USC12K~\citep{zhou2025usc12k}; external scores are literature-reported, our
scores follow the raw three-class protocol below, and SAM2-UNet and Samba report
no USC12K scores. Three effects are worth separating.

\noindent\textbf{Effects that hold.}
{\setlength{\leftmargini}{1.15em}\setlength{\itemsep}{1pt}\setlength{\topsep}{1pt}\setlength{\partopsep}{0pt}\setlength{\parsep}{0pt}
\begin{itemize}
\item \textbf{Type confusion:} CSCS falls from 7.49 to 5.752 ($-23\%$), about
  eight times the seed SD of 0.213 (Supplementary Table~S10).
\item \textbf{Absolute scenes:} Scene-A SOD IoU rises from 74.80 to 83.139 and
  Scene-C COD IoU from 45.73 to 51.487.
\item \textbf{Cost:} 28.80M total parameters against a SAM2 Hiera-L backbone of
  about 224M for USCNet, and a 66.00M PVT-v2-B4 variant reaches 69.77 FG-mIoU and
  5.43 CSCS, so 28.80M is an efficiency setting, not a capacity limit.
\item \textbf{Foreground margin, small:} FG-mIoU moves from 68.455 to 68.983 and
  mIoU from 78.03 to 78.351, about $2.4\times$ the seed standard deviation of
  0.217 (Supplementary Table~S10); this is our weakest claim.
\end{itemize}}

\par\vspace{2pt}\noindent\textbf{Effects that do not.}
{\setlength{\leftmargini}{1.15em}\setlength{\itemsep}{1pt}\setlength{\topsep}{1pt}\setlength{\partopsep}{0pt}\setlength{\parsep}{0pt}
\begin{itemize}
\item \textbf{Single-type COD:} Scene-B COD IoU falls to 69.788 from 74.80,
  which mostly reflects USCNet's larger SAM2 Hiera-L backbone; mAcc nonetheless
  rises from 45.73 to 51.487.
\end{itemize}}\par\vspace{2pt}
Backbone capacity and training budget are unmatched, so no part of the margin
is attributed to the factorization alone. Table~\ref{tab:prediction-paths}
changes the output form and its supervision but not the full architecture, so we
read it as an association, not a causal effect; Supplementary
Tables~S11--S13 re-run the repository's SAM ViT-H build of USCNet (640.07M, not
the paper's Hiera-L) on the same images.

\par\smallskip
\begin{table*}[!t]
\begin{minipage}[t]{0.52\textwidth}
\vspace{0pt}
\paragraph{Type analysis on the conventional benchmarks.}
Tables~\ref{tab:sod-compact}--\ref{tab:joint-compact} evaluate foreground
segmentation only. The table on the right applies the same USC12K weights to the
same benchmarks and reports the type-level quantities of Table~2, which those
tables do not carry. Localization is preserved, but the type decision is
domain-dependent: on COD almost all predicted foreground is typed correctly,
while on SOD a large part of the salient foreground is still labelled
camouflaged, so SOD contributes most of the type error.
\end{minipage}\hfill
\begin{minipage}[t]{0.45\textwidth}
\vspace{0pt}
\centering\footnotesize\rmfamily
\captionof{table}{Type analysis of the unified model on the conventional
benchmarks (\%). Type acc.\ is the share of predicted foreground carrying the
benchmark's true class. CSCS$^{*}$ is one-sided under single-type ground truth.}
\label{tab:transfer}
\setlength{\tabcolsep}{1.2pt}\renewcommand{\arraystretch}{1.2}
\begin{tabular*}{\linewidth}{@{\extracolsep{\fill}}lccccc@{}}
\toprule
\tblhead{Benchmarks} & \tblhead{bIoU$\uparrow$} & \tblhead{IoU-S$\uparrow$} & \tblhead{IoU-C$\uparrow$} & \tblhead{Type acc.$\uparrow$} & \tblhead{CSCS$^{*}\downarrow$}\\
\midrule
SOD (5 sets) & 81.7 & 72.5 & 10.5 & 89.2 & 8.56\\
COD (4 sets) & 70.9 & 3.5 & 70.8 & 95.0 & 6.42\\
\bottomrule
\end{tabular*}
\end{minipage}
\end{table*}

\noindent\begin{minipage}{\columnwidth}
\vspace{0pt}\centering\footnotesize
\captionof{table}{SegDis variants on SOD foreground benchmarks (PVT-v2-B2,
28.80M). SegDis (SOD) is trained on SOD data only, SegDis (Joint) on paired
SOD/COD data, and SegDis (USC12K) is the unified model with no task
identifier. Full results: Supplementary Table~S1. External
rows are source-reported.}
\label{tab:sod-compact}
\setlength{\tabcolsep}{2pt}\renewcommand{\arraystretch}{1.12}
\begin{tabular*}{\linewidth}{@{\extracolsep{\fill}}lccccc@{}}
\toprule
& \multicolumn{2}{c}{\textbf{DUTS-TE}} & \multicolumn{2}{c}{\textbf{ECSSD}}\\
\cmidrule(lr){2-3}\cmidrule(l){4-5}
\makebox[2.35cm][l]{\tblhead{Method}} & \tblhead{$S_\alpha\uparrow$} & \tblhead{MAE$\downarrow$} & \tblhead{$S_\alpha\uparrow$} & \tblhead{MAE$\downarrow$}\\
\midrule
\makebox[2.35cm][l]{\modelcite{F3Net}{wei2020f3net}} & .8880 & .0350 & .9240 & .0330\\
\makebox[2.35cm][l]{\modelcite{VST}{liu2021vst}} & .8960 & .0370 & .9320 & .0330\\
\makebox[2.35cm][l]{\modelcite{ICON-S}{zhuge2022icon}} & .9170 & .0250 & \textbf{.9410} & \textbf{.0230}\\
\makebox[2.35cm][l]{\modelcite{MSFNet}{zhang2021msfnet}} & .8770 & .0340 & .9140 & .0330\\
\makebox[2.35cm][l]{\modelcite{UJSC}{li2021ujsc}} & .8990 & .0320 & .9330 & .0300\\
\makebox[2.35cm][l]{\textbf{SegDis (SOD)}} & .9215 & .0232 & .9395 & .0240\\
\makebox[2.35cm][l]{SegDis (Joint)} & \textbf{.9235} & \textbf{.0230} & .9190 & .0349\\
\makebox[2.35cm][l]{SegDis (USC12K)} & .9061 & .0289 & .9248 & .0316\\
\bottomrule
\end{tabular*}
\end{minipage}\par\vspace{6pt}
\noindent\begin{minipage}{\columnwidth}
\vspace{0pt}\centering\footnotesize
\captionof{table}{SegDis variants on COD foreground benchmarks (PVT-v2-B2,
28.80M). The first SegDis row is trained on COD data only; the second is
trained on paired SOD and COD data. Full results: Supplementary Table~S2. External rows are
source-reported.}
\label{tab:cod-compact}
\setlength{\tabcolsep}{2pt}\renewcommand{\arraystretch}{1.12}
\begin{tabular*}{\linewidth}{@{\extracolsep{\fill}}lccccc@{}}
\toprule
& \multicolumn{2}{c}{\textbf{CAMO}} & \multicolumn{2}{c}{\textbf{COD10K}}\\
\cmidrule(lr){2-3}\cmidrule(l){4-5}
\makebox[2.35cm][l]{\tblhead{Method}} & \tblhead{$S_\alpha\uparrow$} & \tblhead{MAE$\downarrow$} & \tblhead{$S_\alpha\uparrow$} & \tblhead{MAE$\downarrow$}\\
\midrule
\makebox[2.35cm][l]{\modelcite{FSEL-R50}{sun2024fsel}} & .8210 & .0670 & .8300 & .0310\\
\makebox[2.35cm][l]{\modelcite{FSPNet}{huang2023fspnet}} & .8550 & .0500 & .8470 & .0260\\
\makebox[2.35cm][l]{\modelcite{FEDER}{he2023feder}} & .8020 & .0710 & .8200 & .0320\\
\makebox[2.35cm][l]{\modelcite{ZoomNet}{pang2022zoom}} & .8200 & .0660 & .8350 & .0290\\
\makebox[2.35cm][l]{\textbf{SegDis (COD)}} & \textbf{.8664} & \textbf{.0493} & .8673 & \textbf{.0230}\\
\makebox[2.35cm][l]{SegDis (Joint)} & .8651 & .0515 & \textbf{.8693} & .0232\\
\makebox[2.35cm][l]{SegDis (USC12K)} & .8007 & .0802 & .8185 & .0388\\
\bottomrule
\end{tabular*}
\end{minipage}\par\vspace{6pt}

\noindent\begin{minipage}{\columnwidth}
\vspace{0pt}\centering\footnotesize
\captionof{table}{Joint SOD/COD baselines on DUTS-TE and COD10K. The SegDis row
is trained on paired SOD and COD data.
Backbones and full results: Supplementary Table~S6. External rows are
source-reported.}
\label{tab:joint-compact}
\setlength{\tabcolsep}{2pt}\renewcommand{\arraystretch}{1.12}
\begin{tabular*}{\linewidth}{@{\extracolsep{\fill}}lccccc@{}}
\toprule
& \multicolumn{2}{c}{\textbf{DUTS-TE}} & \multicolumn{2}{c}{\textbf{COD10K}}\\
\cmidrule(lr){2-3}\cmidrule(l){4-5}
\makebox[2.35cm][l]{\tblhead{Method}} & \tblhead{$S_\alpha\uparrow$} & \tblhead{MAE$\downarrow$} & \tblhead{$S_\alpha\uparrow$} & \tblhead{MAE$\downarrow$}\\
\midrule
\makebox[2.35cm][l]{\modelcite{UJSC}{li2021ujsc}} & .8990 & .0320 & .8170 & .0350\\
\makebox[2.35cm][l]{\modelcite{SENet}{hao2025senet}} & .9260 & .0220 & .8650 & .0240\\
\makebox[2.35cm][l]{\modelcite{JoNet}{hao2025distribution}} & \textbf{.9310} & \textbf{.0200} & \textbf{.8710} & \textbf{.0230}\\
\makebox[2.35cm][l]{SegDis (Joint)} & .9235 & .0230 & .8693 & .0232\\
\makebox[2.35cm][l]{SegDis (USC12K)} & .9061 & .0289 & .8185 & .0388\\
\bottomrule
\end{tabular*}
\end{minipage}\par\vspace{4pt}

Each table lists two training modes of the same architecture: a model trained
only on that task's data, and a model trained on paired SOD and COD data.
Both use the metric protocol of Supplementary Tables~S4--S6 (raw foreground
probability, no per-image min--max normalisation). These benchmarks evaluate binary foreground discovery only
and complement USC12K's three-class evaluation.
\par\vspace{3pt}

\subsection{CSCS and foreground coverage}
\label{sec:pixel-outcomes}
USC12K~\citep{zhou2025usc12k} introduced CSCS to measure SOD/COD
confusion. Below are its 0--100 forms for SOD-only Scene-A and
COD-only Scene-B, where $n_{a\to b}$ counts true $a$ pixels predicted
as $b$.

\par\smallskip
\noindent\begin{minipage}{\columnwidth}
\centering
\captionsetup{skip=3pt}
\includegraphics[width=\linewidth,trim=0 8bp 0 0,clip]{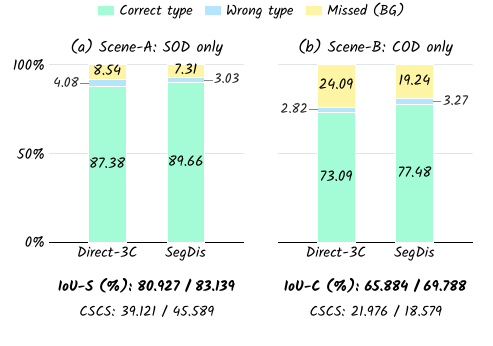}
\captionof{figure}{GT-foreground outcomes on Scenes A and B. Footers give class IoU (\%) and CSCS for Direct-3C / SegDis.}
\label{fig:foreground-outcomes}
\end{minipage}
\par\smallskip

\par\smallskip\noindent\begin{minipage}{\columnwidth}
\textbf{Scene-A.}\hfill
{\small$\displaystyle\mathrm{CSCS}_{A}=50\,
\frac{n_{\mathrm{SOD}\to\mathrm{COD}}}
{n_{\mathrm{SOD}\to\mathrm{COD}}+n_{\mathrm{BG}\to\mathrm{COD}}}$}
\par\smallskip
IoU-S improves (80.927 to 83.139) with fewer GT-foreground swaps and misses.
Yet CSCS rises (39.121 to 45.589): background-to-COD errors fall faster
than SOD-to-COD swaps (74.2\% vs.\ 25.9\%), shrinking its predicted-COD
denominator disproportionately.
\end{minipage}\par\smallskip

\noindent\begin{minipage}{\columnwidth}
\textbf{Scene-B.}\hfill
{\small$\displaystyle\mathrm{CSCS}_{B}=50\,
\frac{n_{\mathrm{COD}\to\mathrm{SOD}}}
{n_{\mathrm{COD}\to\mathrm{SOD}}+n_{\mathrm{BG}\to\mathrm{SOD}}}$}
\par\smallskip
CSCS falls (21.976 to 18.579) despite more GT-foreground swaps
(2.82\% to 3.27\%): increased background-to-SOD errors dilute the swap
fraction. Meanwhile, misses decrease (24.09\% to 19.24\%) and IoU-C
improves (65.884 to 69.788), indicating better foreground recovery.
\end{minipage}\par\smallskip

Figure~\ref{fig:foreground-outcomes} uses all GT foreground as its
denominator, including misses. CSCS instead depends on background false
positives and can move opposite to the GT-foreground swap rate.
It must therefore be read with foreground coverage and class IoUs.
Counts and Scene-C results are in the supplement.

\subsection{FLR and RTA component ablation}
Table~\ref{tab:flr-stages} ablates the two pathway modules in turn, holding
the other complete. Rows F0--F2 add the Master Decoder and the coupled residual
pyramid and Refinement System to the encoder with the complete RTA fixed;
Master decoding alone accounts for 5.098 of the 6.361-point FG-mIoU gain and
residual refinement for the remaining 1.263, so coarse-to-fine foreground
aggregation is the main source of the localization gain and the residual
correction contributes incrementally once the common support exists. Rows
B0--B3 then add the RTA components with the complete FLR fixed: B0 replaces the
factorized path with a three-way head on the same $P_{FG}$, trained with
$\mathcal L_{\mathrm{fg}}$ plus a three-way cross-entropy whose class~0 is
background, B1 adds fixed learned queries with probability composition,
B2 adds relation-conditioned query adaptation, and B3 adds foreground
correction. FG-mIoU rises from 65.840 to 67.749, 68.265, and 68.983, and CSCS
falls from 7.002 to 6.384, 6.257, and 5.752. The two groups meet on the complete
model, since F2 and B3 report the same row.
\par\smallskip
\par\smallskip\noindent\begin{minipage}{\columnwidth}
\centering\fontsize{7.6}{9.0}\selectfont\rmfamily
\captionof{table}{Component ablation of SegDis. E: PVT-v2-B2 encoder.
Rows F add the Master Decoder (MD) and the coupled residual pyramid and
Refinement System (RS) of the FLR pathway with the complete RTA held fixed;
rows B add the RTA components Q, R and C with the complete FLR pathway held
fixed. $\checkmark$ marks a component added at that row, $\dagger$ a component
held complete and not varied, and -- an absent component.}
\label{tab:flr-stages}
\label{tab:fbrc-ablation}
\setlength{\tabcolsep}{.9pt}\renewcommand{\arraystretch}{1.15}
\begin{tabular*}{\linewidth}{@{\extracolsep{\fill}}l*{10}{c}@{}}
\toprule
\tblhead{ID} & \tblhead{E} & \tblhead{MD} & \tblhead{RS} & \tblhead{Q} & \tblhead{R} & \tblhead{C} & \tblhead{FG-mIoU$\uparrow$} & \tblhead{IoU-S$\uparrow$} & \tblhead{IoU-C$\uparrow$} & \tblhead{CSCS$\downarrow$}\\
\midrule
F0 & \textcolor{cvprblue}{$\checkmark$} & -- & -- & \textcolor{cvprblue}{$\dagger$} & \textcolor{cvprblue}{$\dagger$} & \textcolor{cvprblue}{$\dagger$} & 62.622 & 69.233 & 56.010 & 6.113\\
F1 & \textcolor{cvprblue}{$\checkmark$} & \textcolor{cvprblue}{$\checkmark$} & -- & \textcolor{cvprblue}{$\dagger$} & \textcolor{cvprblue}{$\dagger$} & \textcolor{cvprblue}{$\dagger$} & 67.720 & 74.732 & 60.707 & 6.174\\
F2 & \textcolor{cvprblue}{$\checkmark$} & \textcolor{cvprblue}{$\checkmark$} & \textcolor{cvprblue}{$\checkmark$} & \textcolor{cvprblue}{$\dagger$} & \textcolor{cvprblue}{$\dagger$} & \textcolor{cvprblue}{$\dagger$} & 68.983 & 75.663 & 62.304 & 5.752\\
\midrule
B0 & \textcolor{cvprblue}{$\checkmark$} & \textcolor{cvprblue}{$\dagger$} & \textcolor{cvprblue}{$\dagger$} & -- & -- & -- & 65.840 & 74.434 & 57.246 & 7.002\\
B1 & \textcolor{cvprblue}{$\checkmark$} & \textcolor{cvprblue}{$\dagger$} & \textcolor{cvprblue}{$\dagger$} & \textcolor{cvprblue}{$\checkmark$} & -- & -- & 67.749 & 74.831 & 60.668 & 6.384\\
B2 & \textcolor{cvprblue}{$\checkmark$} & \textcolor{cvprblue}{$\dagger$} & \textcolor{cvprblue}{$\dagger$} & \textcolor{cvprblue}{$\checkmark$} & \textcolor{cvprblue}{$\checkmark$} & -- & 68.265 & 75.543 & 60.987 & 6.257\\
B3 & \textcolor{cvprblue}{$\checkmark$} & \textcolor{cvprblue}{$\dagger$} & \textcolor{cvprblue}{$\dagger$} & \textcolor{cvprblue}{$\checkmark$} & \textcolor{cvprblue}{$\checkmark$} & \textcolor{cvprblue}{$\checkmark$} & \textbf{68.983} & \textbf{75.663} & \textbf{62.304} & \textbf{5.752}\\
\bottomrule
\end{tabular*}
\end{minipage}\par\vspace{4pt}

\paragraph{Overall framework.}
Table~\ref{tab:prediction-paths} compares direct three-class heads with
the complete foreground-first pathway. Against the strongest direct head
with the same input sources, $H(F,P_{\mathrm{FG}})$, the factorized path
$\Phi(F,P_{\mathrm{FG}})$ improves FG-mIoU by 1.762 points (67.221 to
68.983); the direct head's readout adds only $0.115$M parameters (S3
projection and classifier) against $0.182$M for RTA, mIoU by 1.238, mAcc by 1.340, IoU-S by 1.693, and IoU-C by
1.833, while reducing CSCS by 0.329. This supports the complete
support-first prediction path over direct three-class prediction: $p_0$
models foreground/background competition, while the conditional type
distribution allocates SOD/COD within foreground. Table~\ref{tab:fbrc-ablation}
reports the cumulative component additions, while the next table compares the
complete path with direct three-class heads.
\par\smallskip
\noindent\begin{minipage}{\columnwidth}
\par\smallskip\noindent\begin{minipage}{\columnwidth}
\centering\fontsize{7.5}{9}\selectfont\rmfamily
\captionof{table}{Framework ablation on USC12K. $H$ denotes a direct
three-class prediction head; $\Phi$ denotes the foreground-first pathway.}
\label{tab:prediction-paths}
\setlength{\tabcolsep}{1.2pt}\renewcommand{\arraystretch}{1.16}
\begin{tabular*}{\linewidth}{@{\extracolsep{\fill}}l*{6}{c}@{}}
\toprule
\tblhead{Model} & \tblhead{FG-mIoU$\uparrow$} & \tblhead{mIoU$\uparrow$} & \tblhead{mAcc$\uparrow$} & \tblhead{IoU-S$\uparrow$} & \tblhead{IoU-C$\uparrow$} & \tblhead{CSCS$\downarrow$}\\
\midrule
$H(F)$ & 65.097 & 75.610 & 82.927 & 72.692 & 57.502 & 6.294\\
$H(P_{\mathrm{FG}})$ & 65.840 & 76.172 & 84.085 & 74.434 & 57.246 & 7.002\\
$H(F,P_{\mathrm{FG}})$ & 67.221 & 77.113 & 84.605 & 73.970 & 60.471 & 6.081\\
\midrule
$\boldsymbol{\Phi}(F,P_{\mathrm{FG}})$ & \textbf{68.983} & \textbf{78.351} & \textbf{85.945} & \textbf{75.663} & \textbf{62.304} & \textbf{5.752}\\
\bottomrule
\end{tabular*}
\end{minipage}\par\smallskip

\end{minipage}\par\smallskip

\subsection{Computational cost}

Table~\ref{tab:runtime} reports the cost comparison at $352\times352$ with
batch size one on a single Quadro RTX 8000. SegDis runs at 49.46\,ms per image ($20.2$\,FPS) with 379\,MiB peak memory and
28.80M parameters, of which RTA contributes 0.182M; the typed output is a single
forward pass, so mixed scenes cost no more. Against the unified baseline this is
22.2$\times$ fewer parameters and 7.2$\times$ less peak memory.
\par\smallskip\noindent\begin{minipage}{\columnwidth}
\centering\footnotesize\rmfamily
\captionof{table}{Costs on Quadro RTX 8000: $352\times352$, batch 1, FP32;
50 warm-up/200 timed passes. Params are total model parameters; FLOPs are
in GFLOPs.}
\label{tab:runtime}
\setlength{\tabcolsep}{1.6pt}\renewcommand{\arraystretch}{1.1}
\newcommand{\ph}[1]{\phantom{#1}}
\begin{tabular*}{\linewidth}{@{\extracolsep{\fill}}ll*{6}{c}@{}}
\toprule
\tblhead{Method} & \tblhead{Backbone} &
\tblhead{\shortstack{Params\\(M)}} &
\tblhead{\shortstack{FLOPs\\(G)}} &
\tblhead{\shortstack{Lat.\\(ms)}} &
\tblhead{FPS} &
\tblhead{\shortstack{Peak.\\(MiB)}}\\
\midrule
\textbf{SegDis} & PVT-v2-B2 & \ph{0}28.80 & \ph{0}34.43 & \ph{0}49.46 & \ph{0}20.22 & \ph{0}379.31\\
ICON-R & ResNet-50 & \ph{0}33.09 & \ph{0}20.91 & \ph{0}28.16 & \ph{0}35.52 & \ph{0}182.28\\
ZoomNet & ResNet-50 & \ph{0}32.38 & \ph{0}85.98 & \ph{0}70.46 & \ph{0}14.19 & \ph{0}337.61\\
CamoFormer-P & PVT-v2-B4 & \ph{0}71.40 & \ph{0}41.75 & \ph{0}93.88 & \ph{0}10.65 & \ph{0}406.58\\
USCNet-SAM & SAM ViT-H & 640.07 & 377.01 & 187.67 & \ph{0}5.33 & 2723.62\\
\bottomrule
\end{tabular*}
\end{minipage}\par\smallskip

\subsection{Qualitative error analysis of type reversals}
Manual inspection found Scene-C type reversals (Figure~\ref{fig:type-ambiguity}),
where the binary foreground-union IoU stays high even though the SOD/COD
assignment on the ground-truth foreground is swapped. These may reflect ambiguity in the SOD/COD distinction itself rather than model
failure: relative area, local contrast, and scene context can influence both
human and model judgment. These are qualitative observations, not a systematic
error distribution; a per-case adjudication study would be needed to quantify
them.
\par\smallskip
\noindent\begin{minipage}{\columnwidth}
\centering
\includegraphics[width=\linewidth]{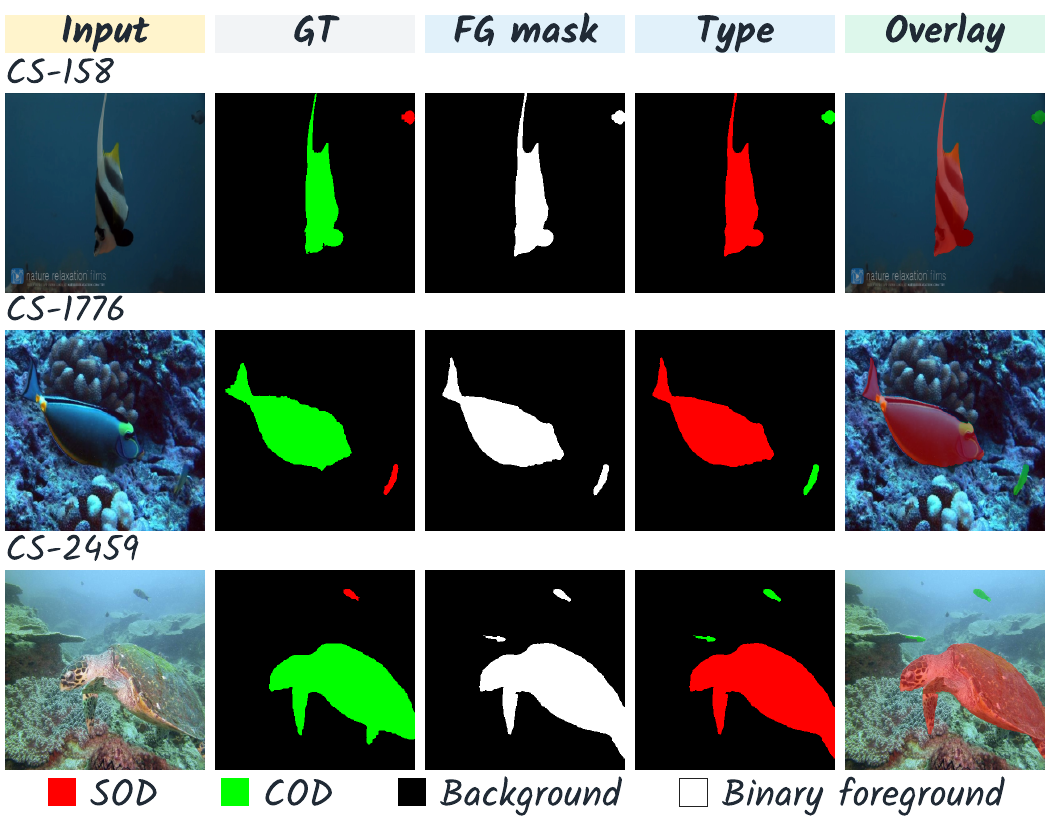}
\captionof{figure}{Scene-C errors: discovered foreground with reversed benchmark types.}
\label{fig:type-ambiguity}
\end{minipage}

\par\smallskip

\par\smallskip

\section{Conclusion}
We present SegDis, a factorized approach to unified foreground discovery and
contextual SOD/COD type assignment, reaching 68.983 FG-mIoU, 78.351 mIoU and
5.752 CSCS. The transfer study shows category-dependent localization, and the
outcome analysis separates missed foreground from type-assignment errors. Beyond
the two benchmarks, one model reporting both support and type can serve as a
front end for monitoring, screening and inspection; broader mixed-scene tests
remain open. Code and trained weights will be released.

\clearpage
\paragraph{LLM Usage Statement.}
OpenAI Codex was used to assist with manuscript polishing and literature search. All technical content, experimental results, and scientific claims were written and verified by the authors.

\begingroup
\footnotesize
\setlength{\bibsep}{0pt}

\endgroup

\fi
\end{document}